\documentclass[10pt,twocolumn,letterpaper]{article}

\usepackage{iccv}
\usepackage{times}
\usepackage{epsfig}
\usepackage{graphicx}
\usepackage{amsmath}
\usepackage{amssymb}

\usepackage{pifont}
\newcommand{\xmark}{\ding{55}}%

\usepackage{colortbl}
\usepackage{booktabs}
\usepackage[table,x11names,dvipsnames,xcdraw]{xcolor}

\usepackage{adjustbox}
\newcommand{\widthscalefive}{0.125}
\newcommand{\widthscalesix}{0.15}

\newcommand{\vecpn}{{\mathbf{p}_n}}
\newcommand{\vecpb}{{\mathbf{p}_b}}
\newcommand{\vecpm}{{\mathbf{p}_m}}

\usepackage{graphicx}               

\usepackage[pagebackref=true,breaklinks=true,letterpaper=true,colorlinks,bookmarks=false]{hyperref}

\iccvfinalcopy

\ificcvfinal\fi
\begin{document}

\title{Motion Deblurring with an Adaptive Network}

\author{Kuldeep Purohit \qquad A. N. Rajagopalan \\
 Indian Institute of Technology Madras, India 
\\
{\tt\small kuldeeppurohit3@gmail.com, raju@ee.iitm.ac.in}
} 

\maketitle

\begin{abstract}
 
In this paper, we address
the problem of dynamic scene deblurring in the presence
of motion blur. Restoration of images affected by severe
blur necessitates a network design with a large receptive field,
which existing networks attempt to achieve through simple increment in the
number of generic convolution layers, kernel-size, or the
scales at which the image is processed. However, increasing the network capacity in this manner comes at the expense
of increase in model size and inference speed, and ignoring
the non-uniform nature of blur. We present a new architecture composed of spatially adaptive residual learning modules that implicitly discover the spatially varying shifts responsible for non-uniform blur in the input image and learn to modulate the filters. This capability is complemented by a self-attentive module which captures non-local relationships among the intermediate features and enhances the receptive field. We then incorporate a spatiotemporal recurrent module in the design to also facilitate efficient video deblurring. Our networks can implicitly model the spatially-varying deblurring process, while dispensing with multi-scale processing
and large filters entirely. Extensive qualitative and quantitative comparisons with prior art on benchmark dynamic
scene deblurring datasets clearly demonstrate the superiority of the proposed networks via reduction in model-size and
significant improvements in accuracy and speed, enabling almost real-time deblurring. 
\end{abstract}

\section{Introduction}

Motion deblurring is a challenging problem in computer vision due to its ill-posed nature. The past decade has witnessed significant advances in deblurring, wherein major efforts have gone into designing priors that are apt for recovering the underlying undistorted image and the camera trajectory \cite{xu2010two,pan2016robust,fergus2006removing,shan2008high,
 cho2009fast,joshi2008psf,krishnan2009fast,krishnan2011blind,xu2010two,
 pan2014deblurring,pan2016blind,vasu2017local,yan2017image, chandramouli2010inferring, paramanand2011depth, paramanand2013non, vijay2013non, rao2014inferring,nimisha2018unsupervised,mohan2019unconstrained,purohit2020region,mohan2021deep,paramanand2014shape,rao2014harnessing, nimisha2017blur,vasu2017local,nimisha2018generating,vasu2018non,purohit2018learning,purohit2019bringing}. An exhaustive survey of uniform blind deblurring algorithms can be found in \cite{lai2016comparative}. Few approaches~\cite{chakrabarti2016neural,schuler2013machine,schuler2016learning} have proposed hybrid algorithms where a Convolutional Neural Network (CNN) estimates the blur kernel, which is then used in an alternative optimization framework for recovering the latent image.

However, these methods have been developed based on a rather strong constraint that the scene is planar and that the blur is governed by only \textit{camera} motion. This precludes commonly occurring blur in most practical settings. Real-world blur arises from various sources including moving objects, camera shake and depth variations, causing different pixels to acquire different motion trajectories. A class of algorithms involve segmentation methods to relax the static and fronto-parallel scene assumption by independently restoring different blurred regions in the scene~\cite{hyun2013dynamic}. However, these methods depend heavily on an accurate segmentation-map. Few methods \cite{sun2015learning,gong2017motion} circumvent the segmentation stage by training CNNs to estimate locally linear blur kernels and feeding them to a non-uniform deblurring algorithm based on patch-level prior. However, they are limited in their capability when it comes to general dynamic scenes.  

Conventional approaches for video deblurring are based
on image deblurring techniques (using priors on the latent sharp frames and the blur kernels) which remove uniform blurs \cite{cai2009blind,zhang2013multi} and non-uniform blur caused by rotational camera motion \cite{li2010generating,cho2012registration,zhang2014multi,zhang2015intra}.  However, these approaches are applicable only under the strong assumption of static scenes and absence of depth-dependent distortions. The work in \cite{wulff2014modeling} proposed a segmentation-based approach to address different blurs in foreground and background regions. Kim et al. \cite{hyun2015generalized} further relaxed the constraint on the scene motion by parameterizing spatially varying blur kernel using optical flow.

With the introduction of labeled realistic motion blur datasets \cite{su2017deep}, deep learning based approaches have been proposed to estimate sharp video frames in an end-to-end manner. Deep Video Deblurring (DVD)~\cite{su2017deep} is the first such work to address generalized video deblurring wherein a neural network accepts a stack of neighboring blurry frames for deblurring. They perform off-line stabilization of the blurred frames before feeding them to the network, which learns to exploit the information from multiple frames to deblur the central frame. Nevertheless, when images are heavily blurred, this method introduces temporal artifacts that become more visible after stabilization. Few methods have also been proposed for burst image deblurring \cite{wieschollek2017learning,aittala2018burst} which utilize number of observations with independent blurs to restore a scene, but are not trained for general video deblurring. Online Video Deblurring (OVD)~\cite{hyun2017online} presents a faster design for video deblurring which does not require frame alignment. It utilizes temporal connections to increase the receptive field of the network. Although OVD can handle large motion blur without adding a computational overhead, it lacks in accuracy and is not real-time.

There are two major limitations shared by prior deblurring works. Firstly, the filters of a generic CNN are spatially invariant (with spatially-uniform receptive field), which is suboptimal to model the process of dynamic scene deblurring and limits their accuracy. Secondly, existing methods achieve high receptive field through networks with a large number of parameters and high computational footprint, making them unsuitable for real-time applications. As the only other work of this kind, \cite{zhang2018dynamic} recently proposed a design composed of multiple CNNs and Recurrent Neural Networks (RNN) to learn spatially varying weights for deblurring. However, their performance is inferior to the state-of-the-art~\cite{tao2018scale} in several aspects. Reaching a trade-off among inference time, accuracy of restoration, and receptive field is a non-trivial task which we address in this paper. We investigate position and motion-aware CNN architecture, which can efficiently handle multiple image segments undergoing motion with different magnitude and direction.

Following recent developments, we adopt an end-to-end learning based approach to directly estimate the restored sharp image. For single image deblurring, we build a fully convolutional architecture equipped with filter-transformation and feature modulation capability suited for the task of motion deblurring. Our design hinges on the fact that motion blur is essentially an aggregation of various spatially varying transformations of the image, and a network that implicitly adapts to the location and direction of such motion, is a better candidate for the restoration task. Next, we address the problem of video deblurring, wherein we extend our single image deblurring network to exploit the redundancy across consecutive frames of a video to guide the process. To this end, we introduce spatio-temporal recurrence at frame and feature-level to efficiently restore sequences of blurred frames.

Our network contains various layers to spatially transform intermediate filters as well as the feature maps. Its advantages over prior art are three-fold: 1. It is fully convolutional and parametrically efficient: deblurring can be achieved with just a single forward pass through a compact network. 2. Its components can be easily introduced into other architectures and trained in an end-to-end manner using conventional loss functions. 3. The transformations estimated by the network are dynamic and hence can be meaningfully interpreted for any test image.

The efficiency of our architecture is demonstrated through comprehensive comparisons with the state-of-the-art on image and video deblurring benchmarks. While a majority of image and video deblurring networks contain $>7$ million parameters, our model achieves superior performance at only a fraction of this size, while being computationally more efficient, resulting in real-time deblurring of images on a single GPU.\\

\section{Proposed Architectures}

An existing technique for accelerating various image processing operations is to down-sample the input image, execute the operation at low resolution, and up-sample the output \cite{chen2016bilateral}. However, this approach discounts the importance of resolution, rendering it unsuitable for image restoration tasks where high-frequency content of the image is of prime importance (deblurring, super-resolution).

Another efficient design is a CNN with a fixed but very large receptive field (comparable to very-high resolution images), e.g. Cascaded dilated network~\cite{chen2017fast}, which was proposed to accelerate various image-to-image tasks. However, simple dilated convolutions are not appropriate for restoration tasks (as shown in \cite{liu2018multi} for image super-resolution). After several layers of dilated filtering, the output only considers a fixed sparse sampling of input locations, resulting in significant loss of information. 

Until recently, the driving force behind performance improvement in deblurring was use of large number of layers, larger filters, and multi-scale processing which gradually increases the ``fixed'' receptive field. Not only is it a suboptimal design, it is also difficult to scale since the effective receptive field of deep CNNs is much smaller than the theoretical one (investigated in \cite{luo2016understanding}).

We claim that a better alternative is to design a convolutional network whose receptive field is adaptive to input image instances. We show that the latter approach is a far better choice due to its task-specific efficacy and utility for computationally limited environments, and it delivers consistent performance across diverse magnitudes of blur. We explain the need for a network with asymmetric filters. Given a 2D image $I$ and a blur kernel $K$, the motion blur process can be formulated as:

\begin{figure}[]
\centering
\includegraphics[width=1.1\linewidth]{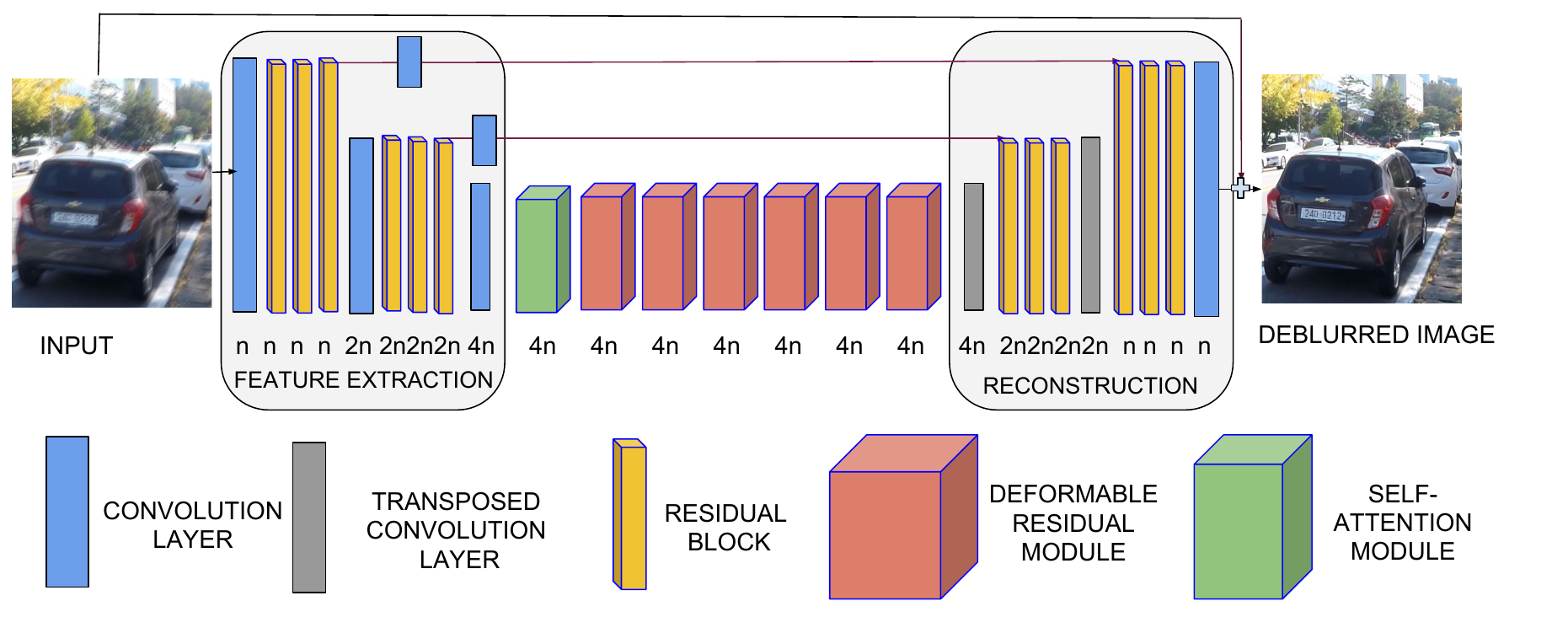}
\caption{The proposed deblurring network and its components.}
\label{fig:architecture}
\end{figure}

\begin{equation}
B[x,y] = \sum_{m,n=-M/2}^{M/2,M/2} K[m,n]I[x-n,y-n],
\end{equation} 
where $B$ is the blurred image, $[x,y]$ represents the pixel coordinates, and $M\times M$ is the size of the blur kernel. At any given location $[x,y]$, the sharp intensity can be represented as  
\begin{equation}
I[x,y] =\frac{B[x,y]}{K[0,0]}-\frac{\sum_{m,n=-M/2}^{M/2,M/2}K[m,n]I[x-n,y-n]}{K[0,0]},
\end{equation} 
which is a 2D infinite impulse response (IIR) model. Recursive expansion of the second term would eventually lead to an expression which contains values from only the blurred image and the kernel as

\begin{small}
    \begin{equation} 
\hspace{-0.15cm}
\begin{aligned}
I[x,y] =\frac{B[x,y]}{K[0,0]}-\sum_{m,n=-M/2}^{M/2,M/2}\frac{K[m,n]B[x-m,y-n]}{K[0,0]^2}+ \\\frac{\sum_{m,n=-M/2}^{M/2,M/2}\sum_{i,j=-M/2}^{M/2,M/2}K[m,n]K[i,j]I[x-n-i,y-n-j]}{K[0,0]^2}
\end{aligned}
\label{iir}
\end{equation} 
\end{small}

The dependence of $I[x,y]$ on a large number of locations in $B$ shows that the deconvolution process requires infinite signal information. If we assume that the boundary of the image is zero, eq. \ref{iir} is equivalent to applying an inverse filter to $B$. As visualized in \cite{zhang2018dynamic}, the non-zero region of such an inverse deblurring filter is typically much larger than the blur kernel. Thus, if we use a CNN to model the process, a large receptive field should be considered to cover the pixel positions that are necessary for deblurring. Eq. \ref{iir} also shows that only a few coefficients (which are $K[m,n]$ for $m,n \in [-M/2,M/2]$) need to be estimated by the deblurring model, provided we can find an appropriate operation to cover a large enough receptive field.

For this theoretical analysis, we will temporarily assume that the motion blur kernel $K$ is linear (assumption used in few prior deblurring works \cite{sun2015learning,gong2017motion}). Now, consider an  image $B$ which is affected by motion blur in the horizontal direction (without loss of generality), implying $K[m,n]=0$ for $m \neq 0$ (non-zero values present only in the middle row of the kernel). For such a case, eq. \ref{iir} translates to
\begin{equation} 
\hspace{-0.5cm}
\begin{aligned}
I[x,y] =\frac{B[x,y]}{K[0,0]}-\sum_{n=1}^{M}\frac{K[o,n]B[x,y-n]}{K[0,0]^2}+ \\\frac{\sum_{n=1}^{M}\sum_{j=1}^{M}K[0,n]K[0,j]I[x,y-n-j]}{K[0,0]^2}=...
\end{aligned}
\label{iir2}
\end{equation} 

It can be seen that for this case, $I[x,y]$ can be expressed as a function of only one row of pixels in the blurred image $B$, which implies that for a horizontal blur kernel, the deblurring filter is also purely horizontal. We use this observation to state a hypothesis that holds for any motion blur kernel: ``Deblurrig filters are directional/asymmetric in shape''. This is because motion blur kernels are known to be inherently directional. Such an operation can be efficiently learnt by a CNN with adaptive and asymmetric filters and this forms the basis for our work.

Inspired by the success of deblurring works that utilize networks composed of residual blocks to directly regress to the sharp image \cite{nimisha2017blur,nah2017deep,tao2018scale}, we build our network over a residual encoder-decoder structure. Such a structure was adopted in Scale Recurrent Network (SRN)~\cite{tao2018scale}, which is the current state-of-the-art in deblurring. We differentiate our design from SRN in terms of compactness and computational footprint. While SRN is composed of $5\times5$ conv filters, we employ only $3\times3$ filters for economy. Unlike \cite{tao2018scale}, our single image deblurring network does not contain recurrent units, and most importantly, our approach does not involve multi-scale processing; the input image undergoes only a single pass through the network. Understandably, these changes can drastically reduce the inference time of the network and also decrease the model's representational capacity and receptive field in comparison to SRN, with potential for significant reduction in the deblurring performance. In what follows, we describe our proposed architecture which matches the efficiency of above network while significantly improving representational capacity and performance.

In our proposed Spatially-Adaptive Residual Network (SARN), the encoder sub-network progressively transforms the input image into feature maps with smaller spatial size and more channels. Our spatially adaptive modules (Deformable Residual Module (DRM) and Spatial Attention (SA) module) operate on the output of the encoder, where the spatial resolution of features is the smallest which leads to minimum additional computations. The resulting features are fed to the Decoder, wherein it is passed through a series of Res-Blocks and deconvolution layers to reconstruct the output image. A schematic of the proposed architecture is shown in Fig. \ref{fig:architecture}, where $n$ (=$32$) represents the number of channels in the first feature map. Next, we describe the proposed modules in detail.

\begin{figure}[]
\centering
\includegraphics[width=\linewidth]{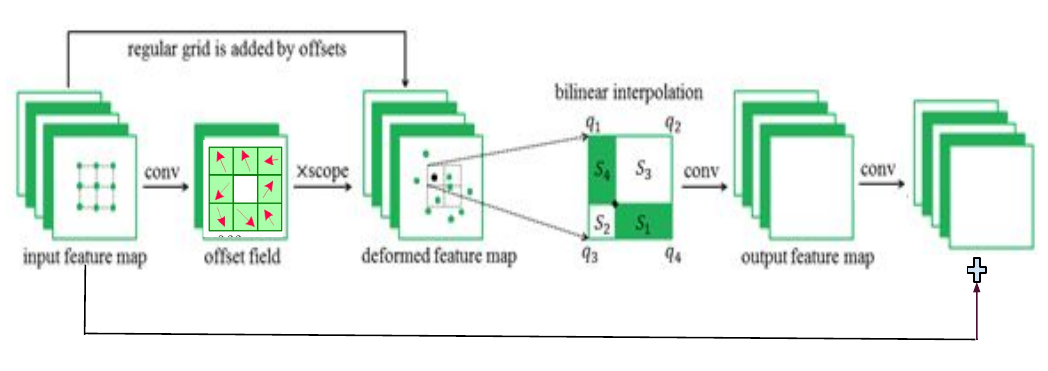}
\caption{Schematic of our deformable residual module.}
\label{fig:deformableblock}
\end{figure}

\subsection{Deformable Residual Module (DRM)}
CNNs operate on fixed locations in a regular grid which limits their ability to model unknown geometric transformations. Spatial Transform Networks (STN) \cite{jaderberg2015spatial} introduced spatial transformation learning into CNNs, wherein an image-dependent global parametric transformation is estimated and applied on the feature map. However, such warping is computationally expensive and the transformation is considered to be global across the whole image, which is not the case for motion in dynamic and 3D scenes where different regions are affected by different magnitude and direction of motion. Hence, we adopt deformable convolutions~\cite{dai2017deformable}, which enable local transformation learning in an efficient manner. Unlike regular convolutional layers, the deformable convolution\cite{dai2017deformable} also learns to estimate the shapes of convolution filters conditioned on an input feature map. While maintaining filter weights invariant to the input, a deformable convolution layer first learns a dense offset map from the input, and then applies it to the regular feature map for re-sampling.

As shown in Fig. \ref{fig:deformableblock}, our DRM contains the additional capability to a learn positions of the sampling grid used in the convolution. A regular convolution layer is present to estimate the features and another convolution layer to estimate 2D filter offsets for each spatial location. These channels (feature-maps containing red-arrows in Fig. \ref{fig:deformableblock}) represent the estimated 2D offset of each input. The 2D offsets are encoded in the channel dimension i.e., convolution layer of $k \times k$ filters is paired with offset predicting convolution layer of $2 k^2$ channels. These offsets determine the shifting of the $k^2$ filter locations along horizontal and vertical axes. As a result, the regular convolution filter operates on an irregular grid of pixels. Since the offsets can be fractional, bilinear interpolation is used to sample from the input feature map. All the parts of our network are trainable end-to-end, since bilinear sampling and the grid generation of the warping module are both differentiable \cite{paszke2017automatic}. The offsets are initialized to $0$. Finally, the additive link grants the benefits of reusing common features with low redundancy.

The convolution operator slides a filter or kernel over the input feature map $\mathbf{X}$ to produce output feature map $\mathbf{Y.}$ For each sliding position $\vecpb$, a regular convolution with filter weights $\mathbf{W},$ bias term $\mathbf{b}$ and stride 1 can be formulated as
$$\mathbf{Y}=\mathbf{W}*\mathbf{X}+\mathbf{b}$$
\begin{equation}
\label{eqn:rconv}
y_\vecpb=\sum_c{\sum_{\vecpn\in\mathcal{R}}{w_{c,n}\cdot x_{c,\vecpb+\vecpn}+b}}
\end{equation}
where $c$ is the index of input channel, $\vecpb$ is the base position of the convolution, $n=1,\ldots,N$ with $N=|{\mathcal{R}|}$ and $\vecpn\in\mathcal{R}$ enumerates the locations in the regular grid $\mathcal{R}$. The center of $\mathcal{R}$ is denoted as $\vecpm$ which is always equal to $(0,0)$, under the assumption that both height and width of the kernel are odd numbers. This assumption is suitable for most CNNs. $m$ is the index of the central location in $\mathcal{R}$.

The deformable convolution augments all the sampling locations with learned offsets $\{\Delta\mathbf{p}_n|n=1,\ldots,N\}$. Each offset has a horizontal component and a vertical component. Totally $2N$ offset parameters are required to be learnt for each sliding position. Equation (\ref{eqn:rconv}) then becomes
\begin{equation}
\label{eqn:dconv}
y_\vecpb=\sum_{\vecpn\in\mathcal{R}}{w_n\cdot x_{\mathbf{H}(\vecpn)}+b}
\end{equation}
where $\mathbf{H}(\vecpn)=\vecpb+\vecpn+\Delta\vecpn$ is the learned sampling position on input feature map. The input channel $c$ in (\ref{eqn:rconv}) is omitted in (\ref{eqn:dconv}) for notational clarity, because the same operation is applied in every channel.

The receptive field and the spatial sampling locations are adapted according to the scale, shape, and location of the degradation. Presence of a cascade of DRMs imparts higher accuracy to the network while delivering higher parameter efficiency than the state-of-the-art deblurring approaches. Although the focus of our work is a compact network design, it also provides an effective way to further increase the network capacity since replacing normal Res-Blocks with DRMs is much more efficient than going deeper or wider. In our final network, $6$ DRMs are present in the mid-level of the network.

 \subsection{Video Deblurring through Spatio-temporal recurrence}

A natural extension to single image deblurring is video deblurring. However, video deblurring is a more structured problem as it can utilize information distributed across multiple observations to mitigate the ill-posedness of deblurring. Existing learning-based approaches \cite{su2017deep,hyun2017online} have proposed generic encoder-decoder architectures to aggregate information from neighboring frames. At each time step, DVD \cite{su2017deep} accepts a stack of neighboring blurred frames as input to network, while OVD \cite{hyun2017online} accepts intermediate features extracted from past frames.
 
We present an effective technique which elegantly extends our efficient single image deblurring design to restore a sequence of blurred frames. The proposed network encourages recurrent information propagation along the temporal direction  at feature-level as well as frame-level to achieve temporal consistency and improve restoration quality. For feature propagation, our network employs Convolutional Long-Short Term Memory (LSTM) modules \cite{xingjian2015convolutional} which are known to efficiently process spatio-temporal data and perform gated feature propagation. The process can be expressed as
\begin{equation}
   \begin{aligned}
      \mathbf{f}^i &= \mathbf{Net}_{E}(\mathbf{B}^i,\mathbf{I}^{i-1}), \\
      \mathbf{h}^i,\mathbf{g}^i &= \mathbf{ConvLSTM}(\mathbf{h}^{i-1},\mathbf{f}^i; \theta_{LSTM}),\\
      \mathbf{I}^i &= \mathbf{Net}_{D}(\mathbf{g}^i; \theta_D),
   \end{aligned}\label{eq:edresblock}
\end{equation}
where $i$ represents frame index, $\mathbf{Net}_{D}$ is the decoder part of our network with parameters $\theta_D$ and $\mathbf{Net}_{E}$ is the portion before the decoder. $\theta_{LSTM}$ is the set of parameters in ConvLSTM. The hidden state $h^i$ contains useful information about intermediate results and blur patterns, which are passed to the network processing the next frame, thus assisting it in sharp feature aggregation.

 Unlike \cite{su2017deep,hyun2017online}, our framework also employs recurrence at frame level wherein previously deblurred estimates are provided at the input to the network that processes subsequent frames. This naturally encourages temporally consistent results by allowing it to assimilate a large number of previous frames without increased computational demands. Our network accepts 5 frames at each time-step (early fusion), of which 2 frames are deblurred estimates from past and 2 are blurred frames from future. As discussed in \cite{caballero2017real}, such early fusion allows the initial layers to assimilate complementary information from neighboring frames and improves restoration quality.

\begin{figure*}[htb]
	\scriptsize
	\centering
			\begin{tabular}{ccccccc}
								\includegraphics[width=0.12\textwidth]{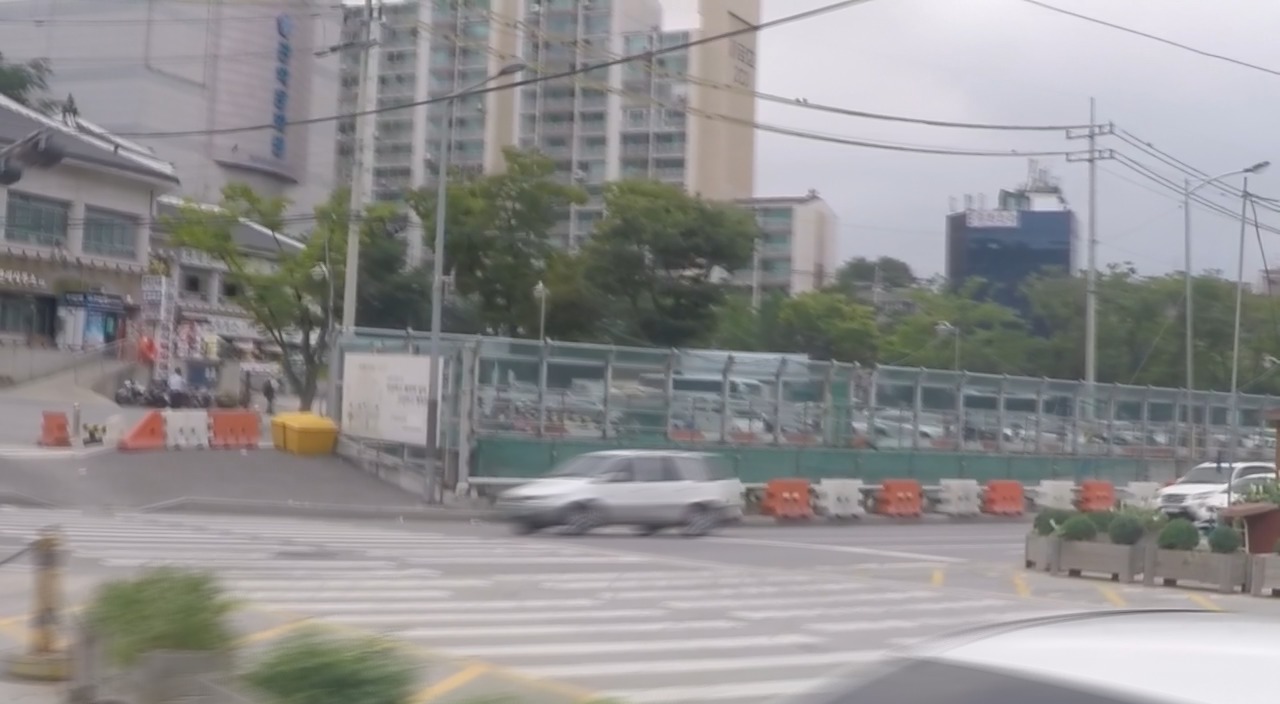} & \includegraphics[width=\widthscalefive \textwidth]{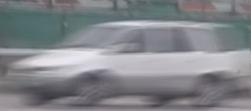} & 
				\includegraphics[bb=499 178 749 288,clip=True,width=\widthscalefive \textwidth]{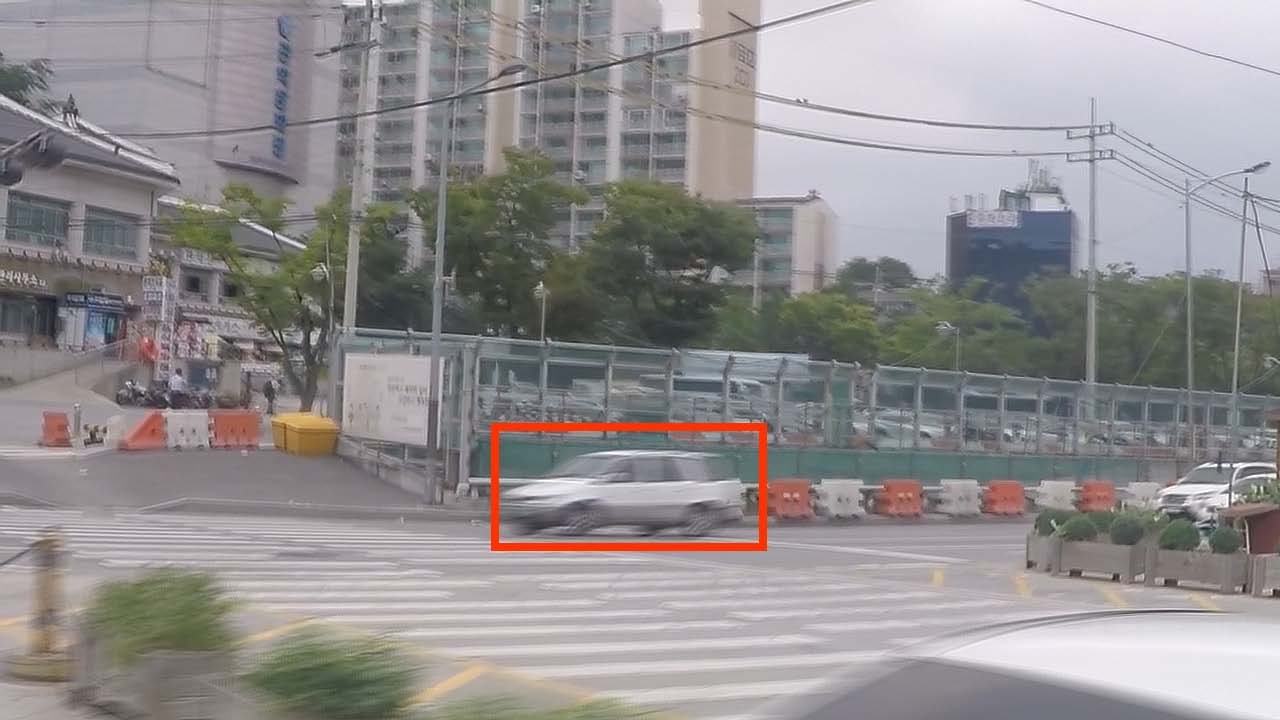} & 
				\includegraphics[width=\widthscalefive \textwidth]{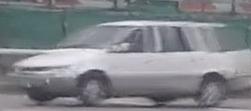} & 
				\includegraphics[width=\widthscalefive \textwidth]{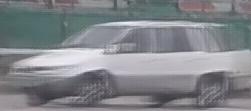} & 
								\includegraphics[bb=490 170 740 280,clip=True,width=\widthscalefive \textwidth]{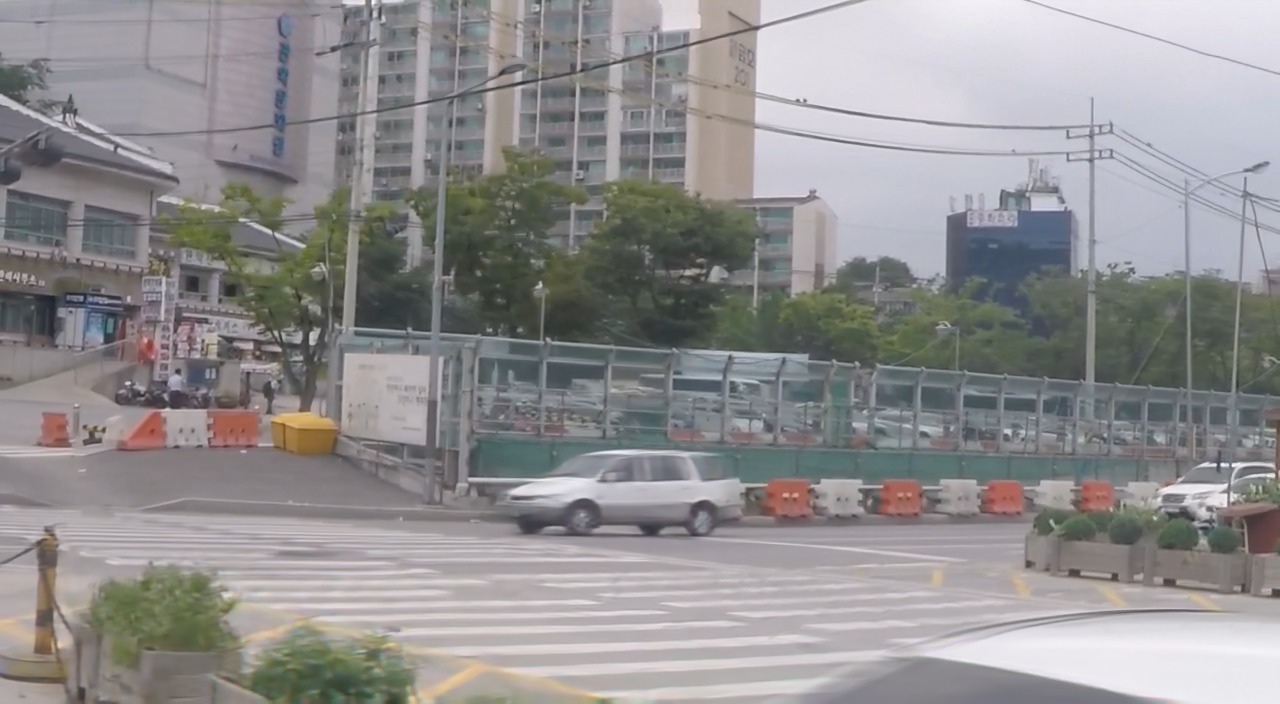} & 
				\includegraphics[width=\widthscalefive \textwidth]{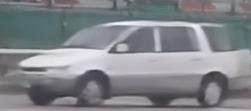}
				\\
				\includegraphics[width=0.12\textwidth]{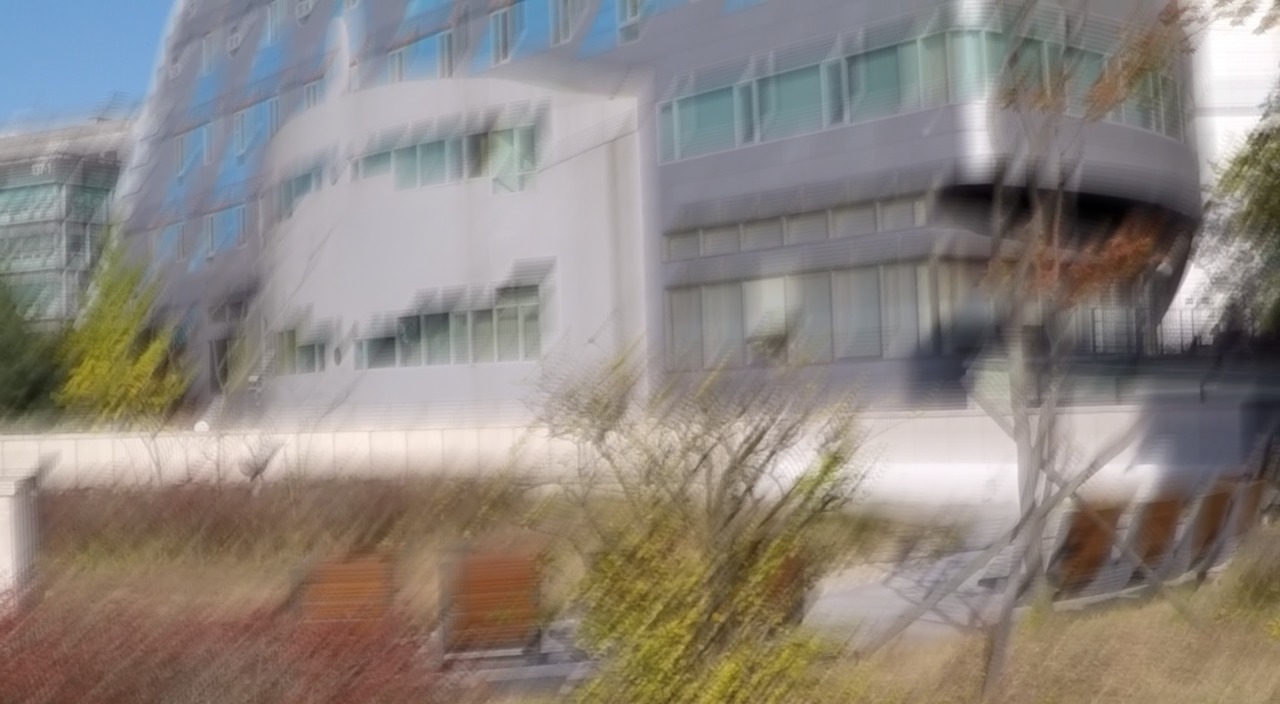}
				&
				\includegraphics[bb=480 1 880 200,clip=True,width=\widthscalefive \textwidth]{deblurring/000045_blurred.jpg} & 
				\includegraphics[bb=480 1 880 200,clip=True,width=\widthscalefive \textwidth]{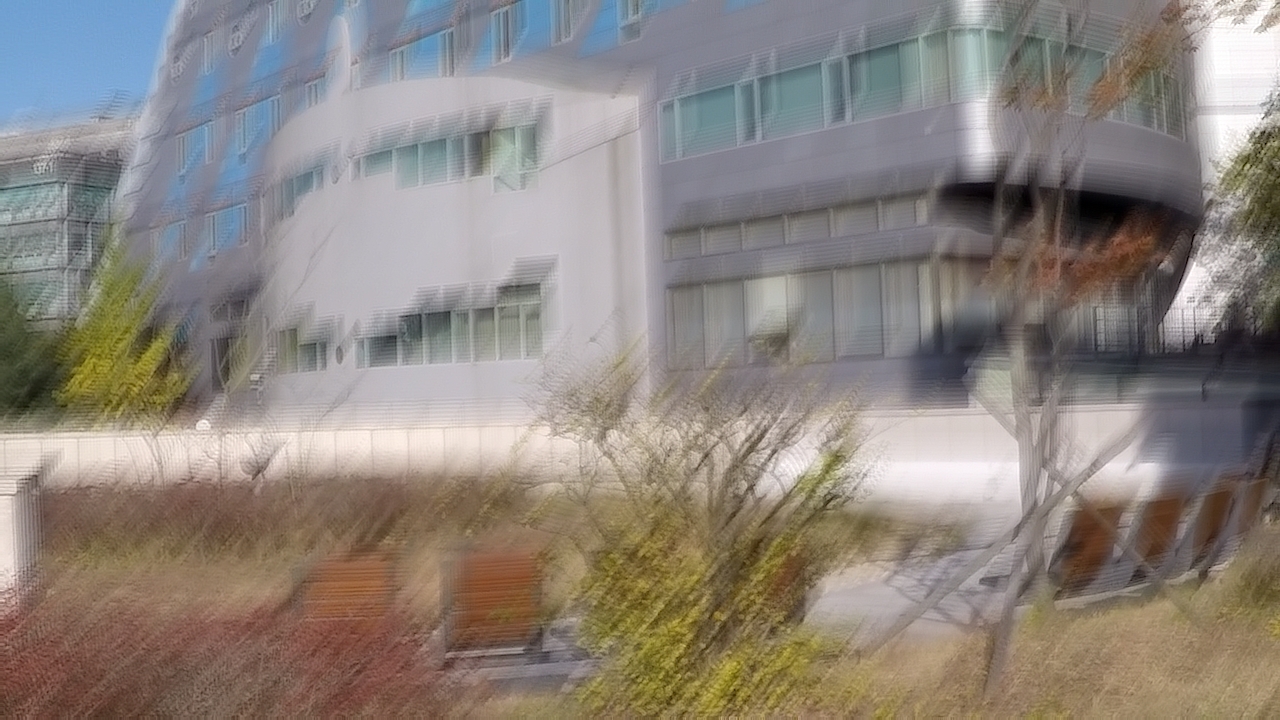} & 
				\includegraphics[bb=480 1 880 200,clip=True,width=\widthscalefive \textwidth]{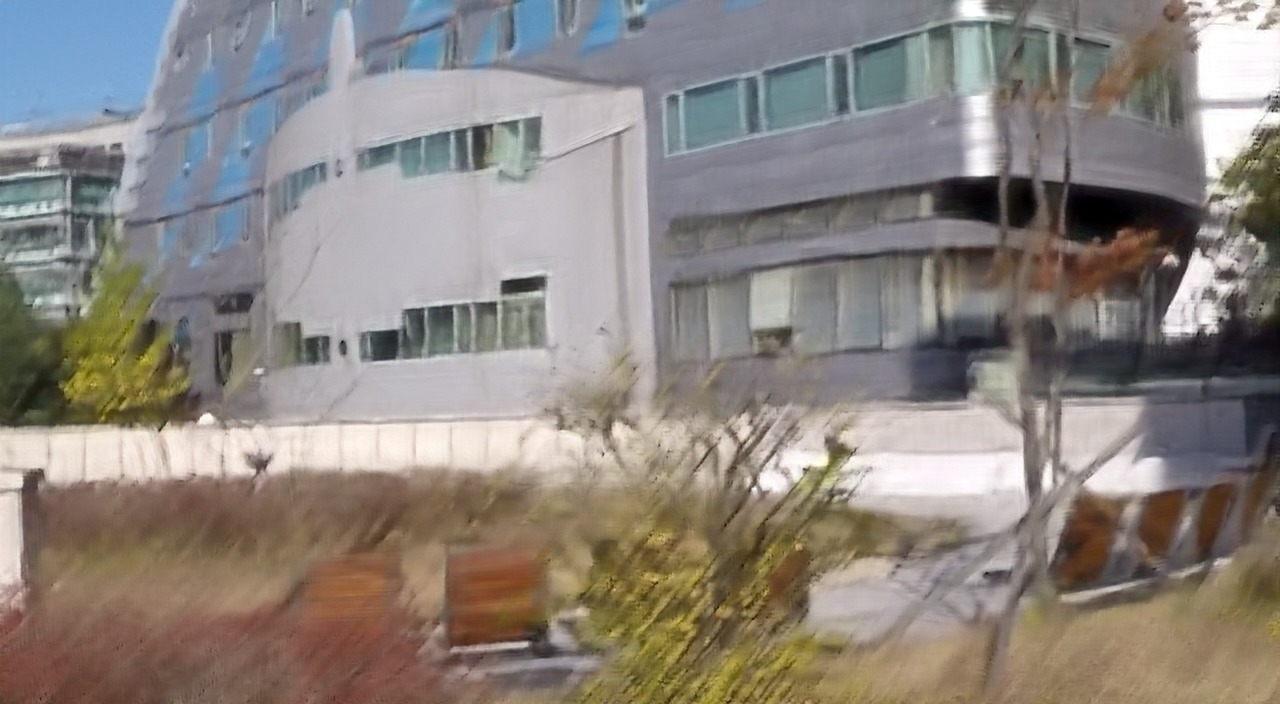} & 
				\includegraphics[bb=480 1 880 200,clip=True,width=\widthscalefive \textwidth]{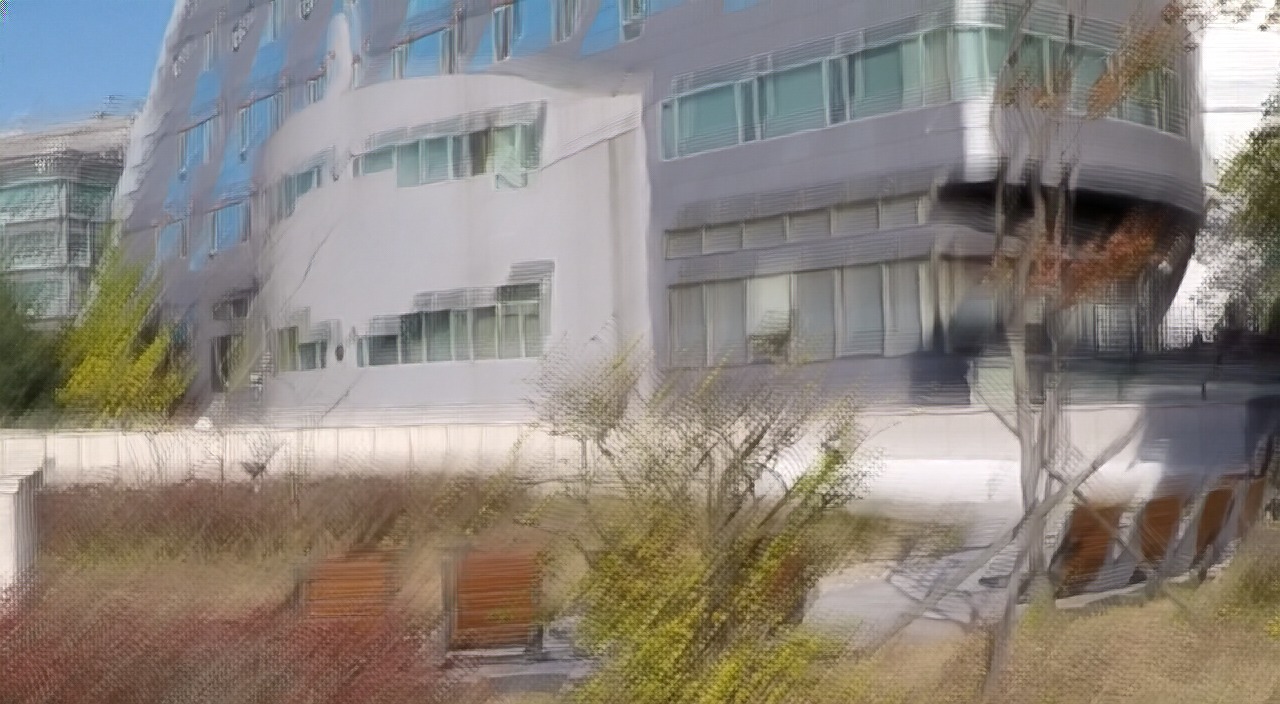} & 
								\includegraphics[bb=480 1 880 200 280,clip=True,width=\widthscalefive \textwidth]{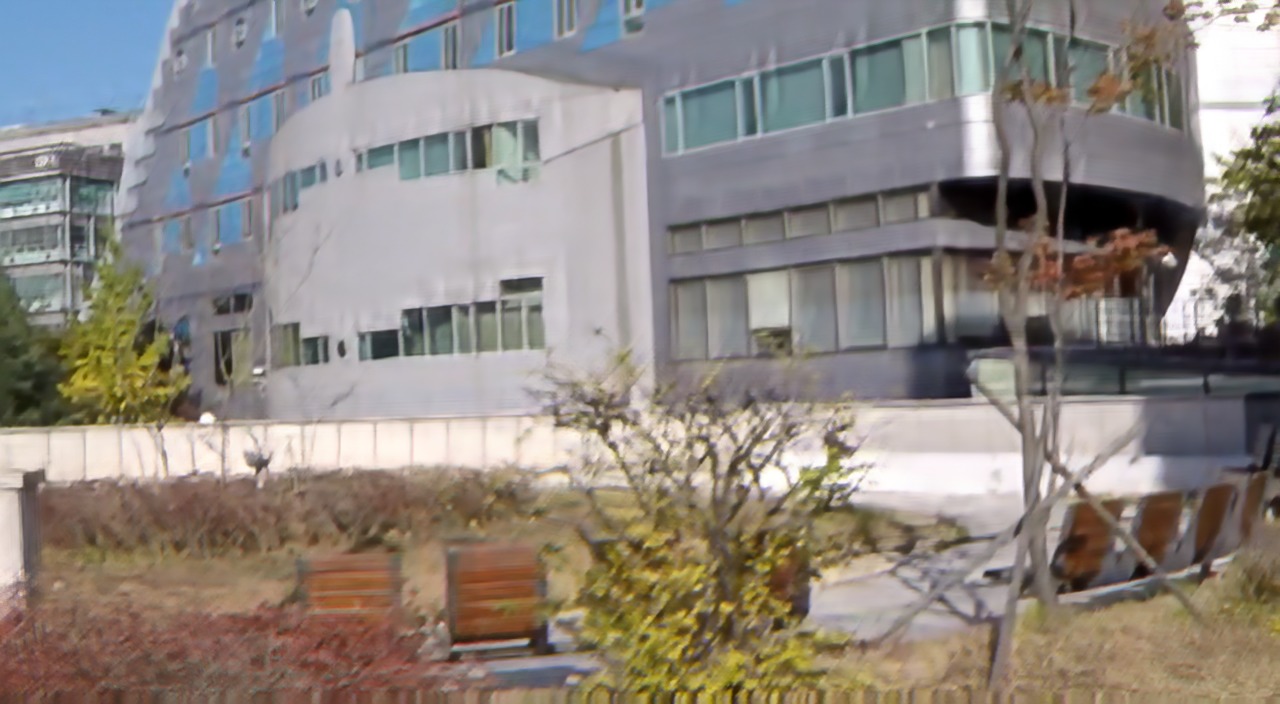} & 
				\includegraphics[bb=480 1 880 200,clip=True,width=\widthscalefive \textwidth]{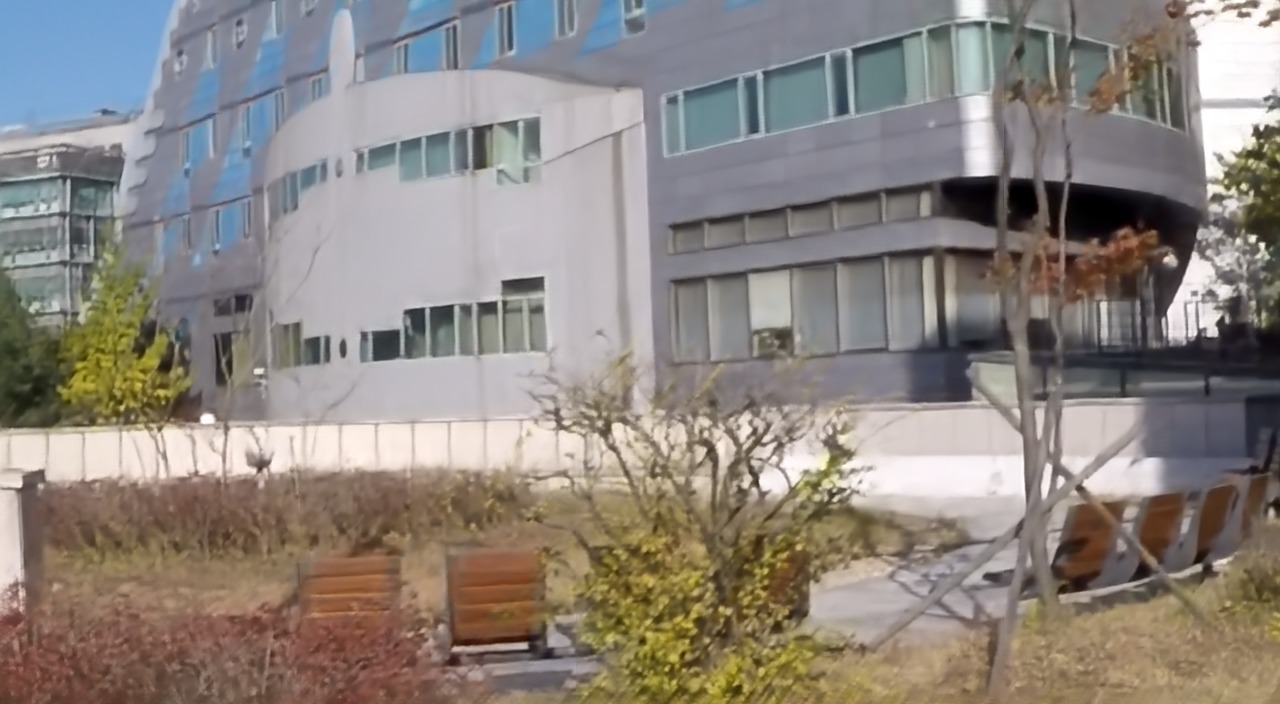}				\vspace{-3mm}
\\
\\
				\includegraphics[width=0.12\textwidth]{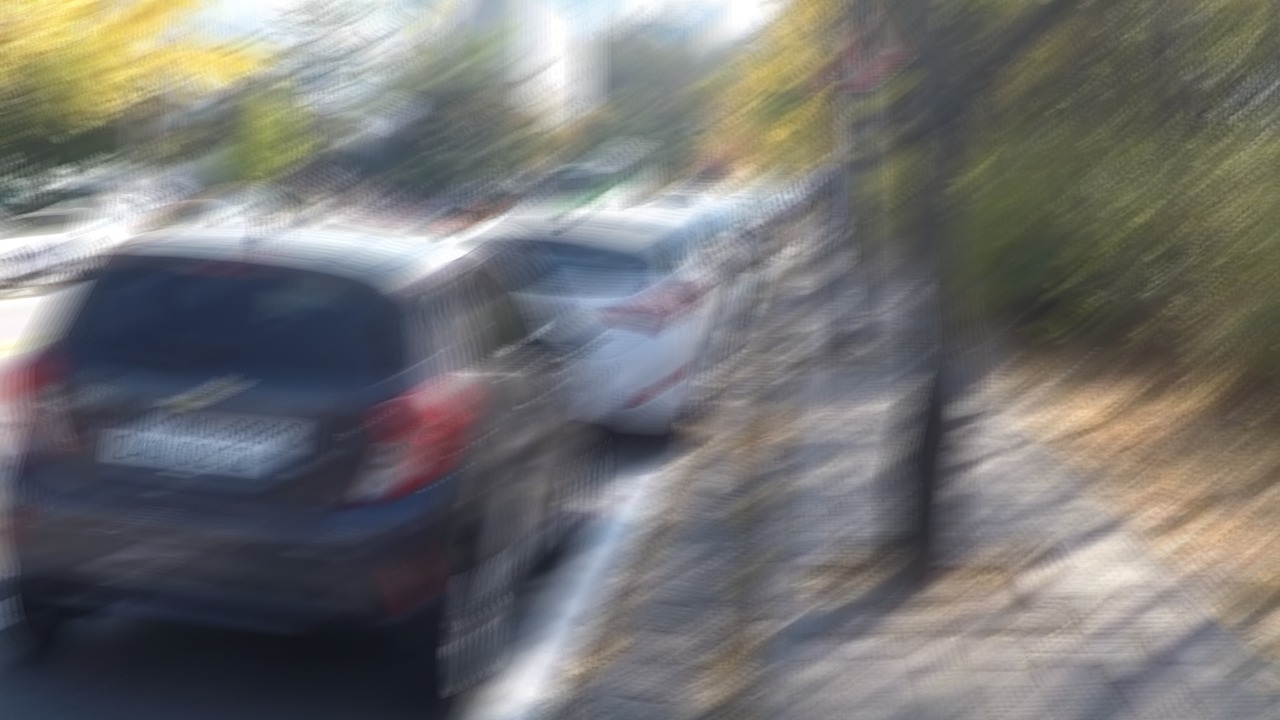}
				&
				\includegraphics[bb=80 200 380 370,clip=True,width=\widthscalefive \textwidth]{deblurring/000006_blurred.jpg} & 
				\includegraphics[bb=80 200 380 370,clip=True,width=\widthscalefive \textwidth]{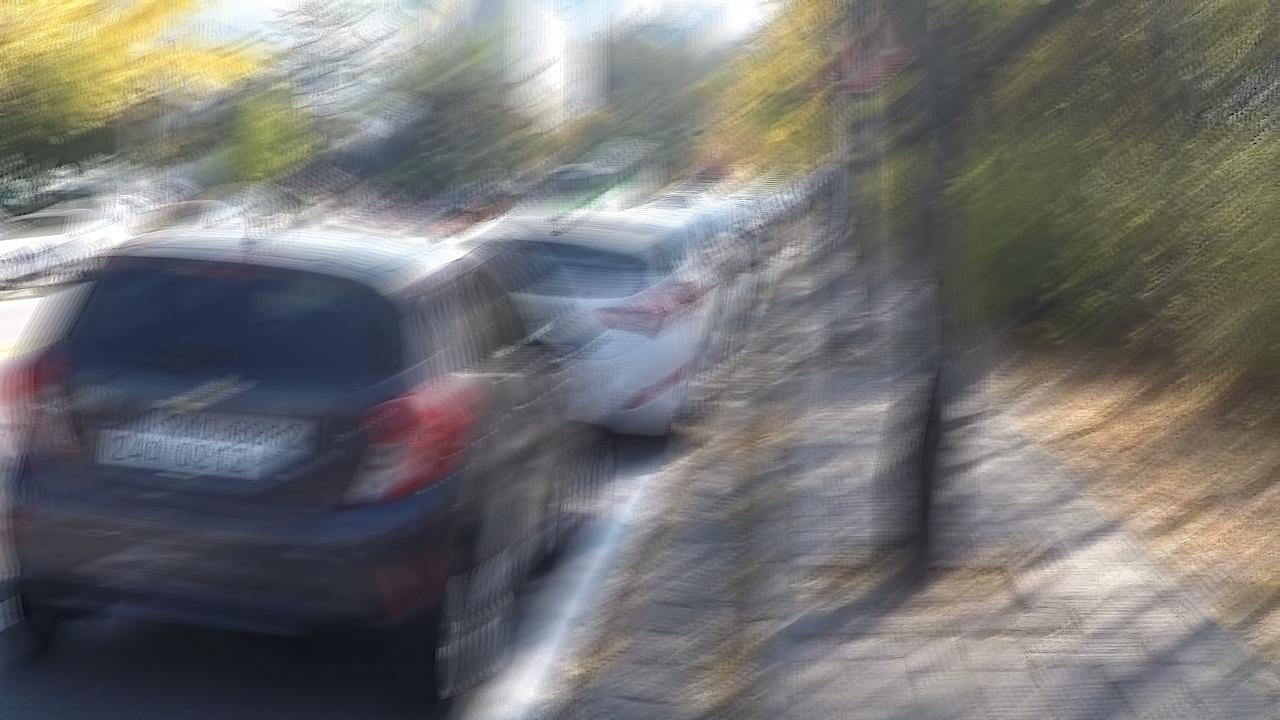} & 
				\includegraphics[bb=80 200 380 370,clip=True,width=\widthscalefive \textwidth]{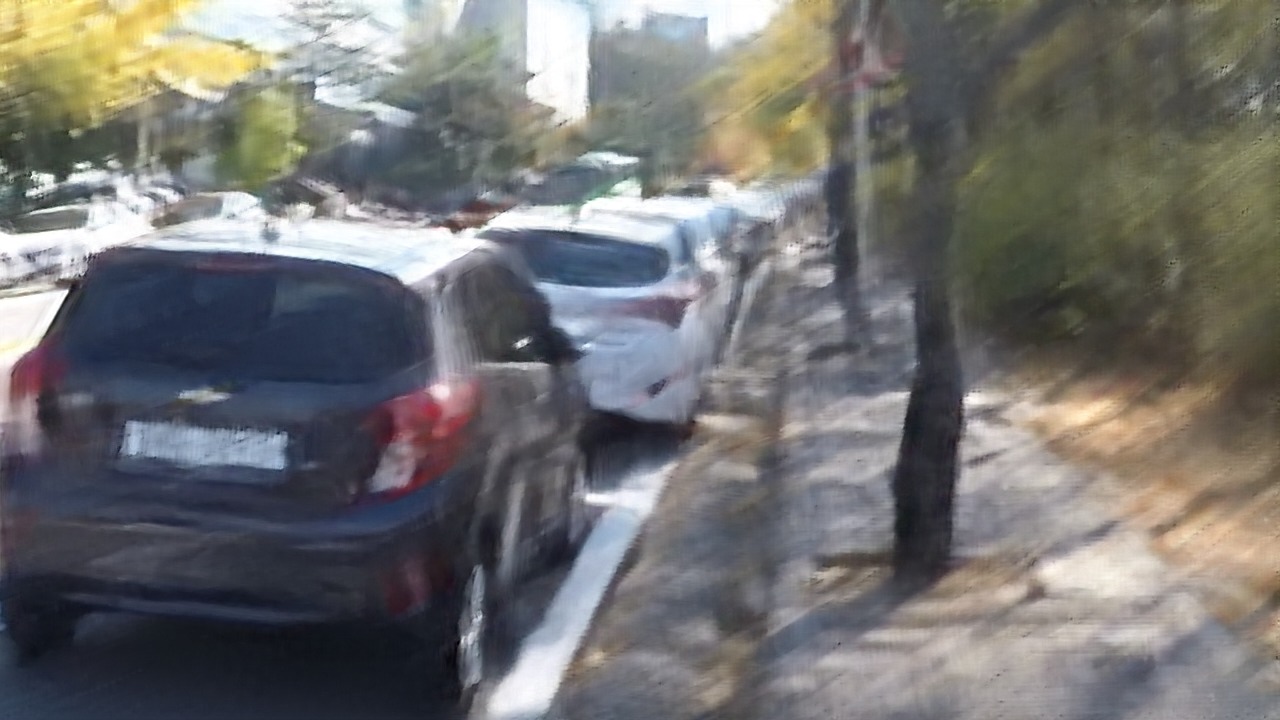} & 
				\includegraphics[bb=80 200 380 370,clip=True,width=\widthscalefive \textwidth]{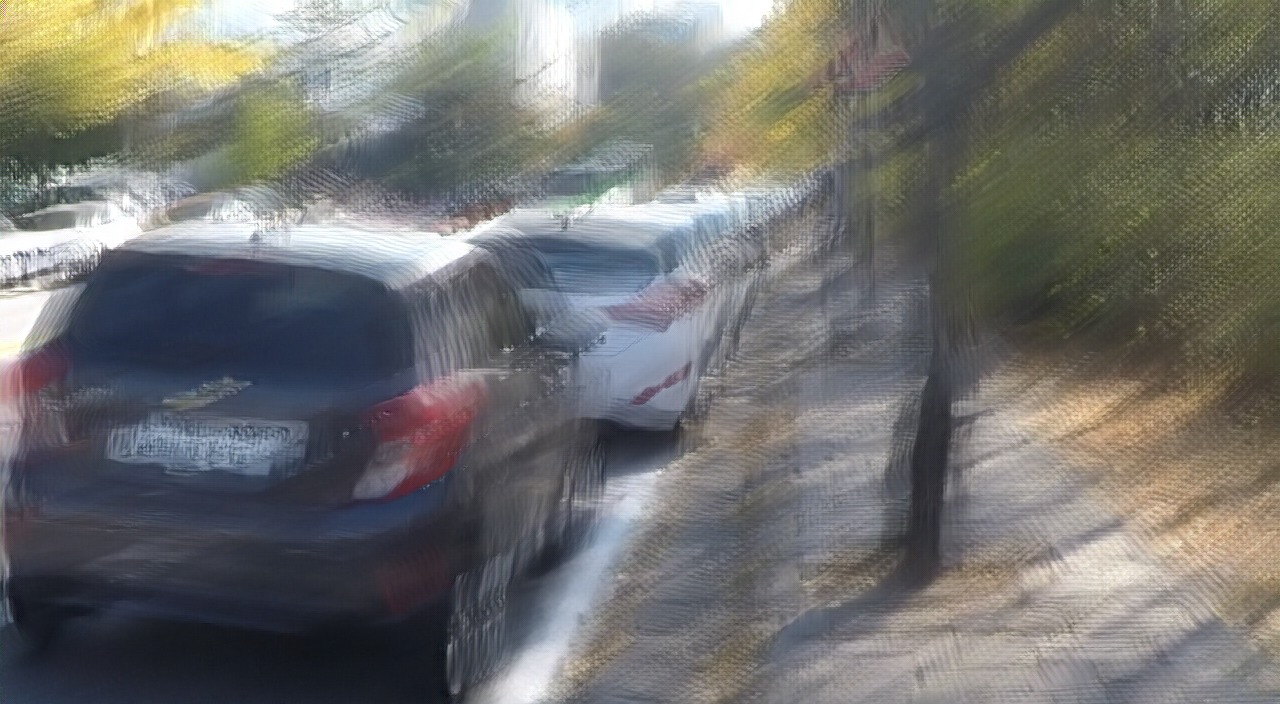} & 
								\includegraphics[bb=80 200 380 370,clip=True,width=\widthscalefive \textwidth]{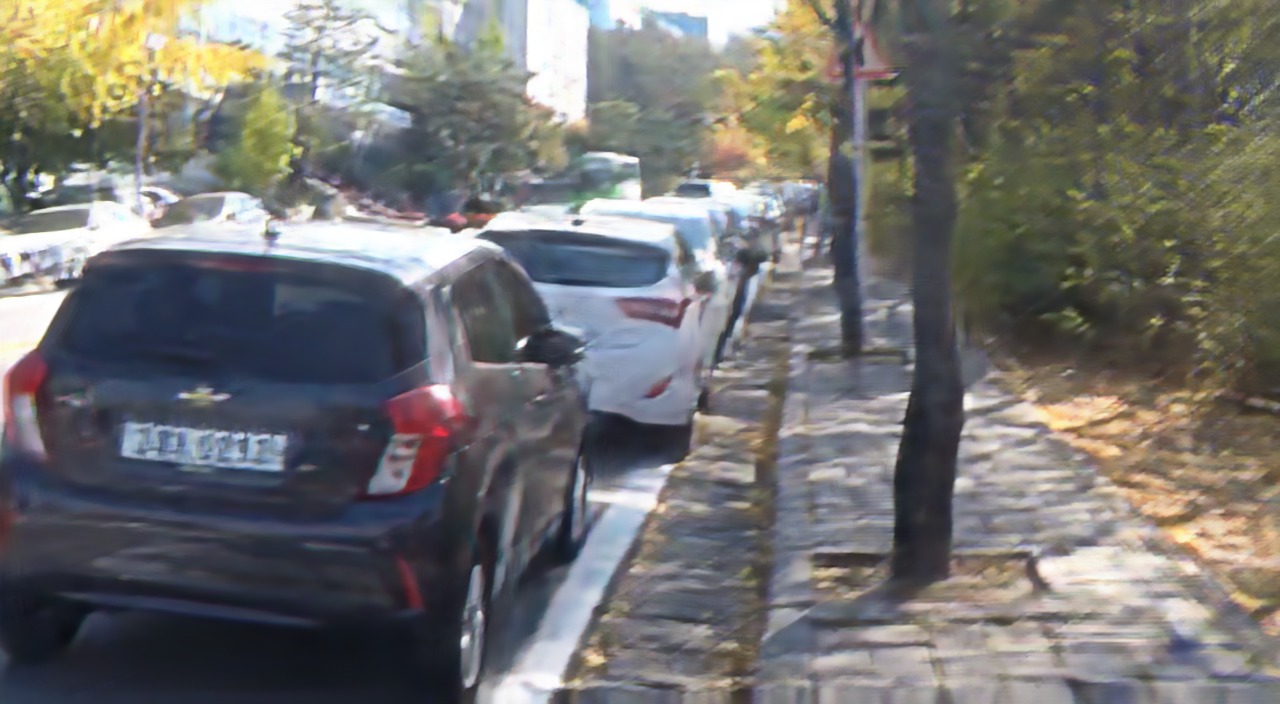} & 
				\includegraphics[bb=80 200 380 370,clip=True,width=\widthscalefive \textwidth]{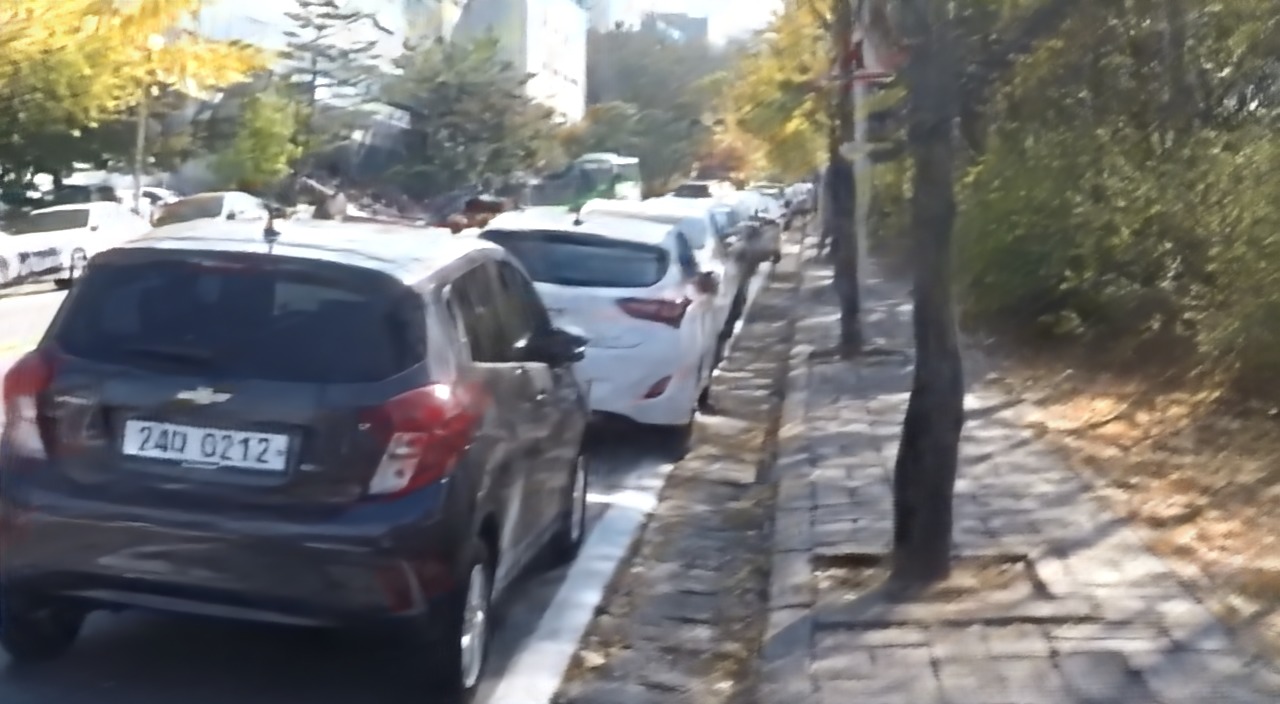}				\vspace{-3mm}

				\\ 
				\\ 
								(a) Blurred Image&

				(b) Blurred patch& 
				(c) Whyte \etal~\cite{whyte2012non} & 
				(d) Nah \etal~\cite{nah2017deep} & 
				(e) DelurGAN~\cite{kupyn2017deblurgan}& 
				(f) SRN~\cite{tao2018scale}& 
				(g) Ours
				\\
	\end{tabular}
	\vspace{-0em}
	\caption{Visual comparisons of deblurring results on test images from the GoPro dataset~\cite{nah2017deep}. Key blurred patches are shown in (b), while zoomed-in patches from the deblurred results are shown in (c)-(g). (best viewed in high resolution).}
\label{fig:dynamic}
	\vspace{-0em}
\end{figure*}

\begin{figure*}[!htb]
	\scriptsize
	\centering
			\begin{tabular}{cccccc}
				\includegraphics[width=0.12\textwidth]{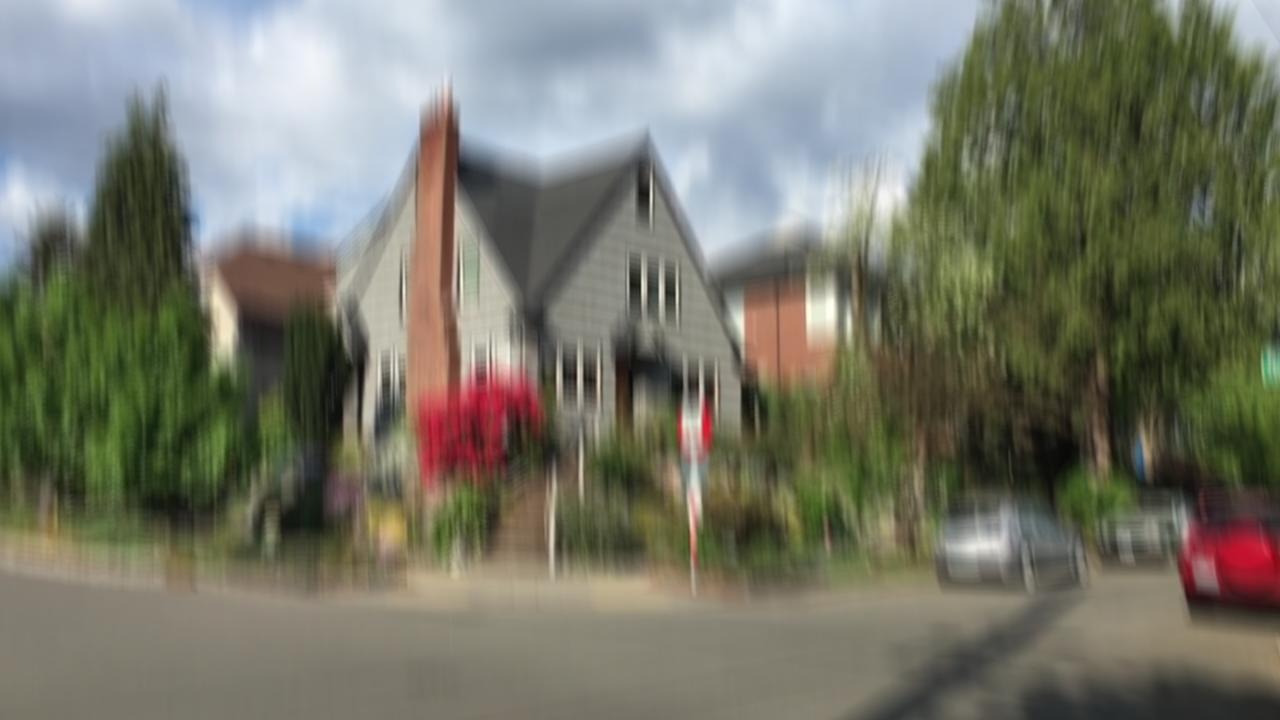}
				&
				\includegraphics[bb=500 300 800 450,clip=True,width=\widthscalesix \textwidth]{video_real/00082_blur.jpg} & 
				\includegraphics[bb=500 300 800 450,clip=True,width=\widthscalesix \textwidth]{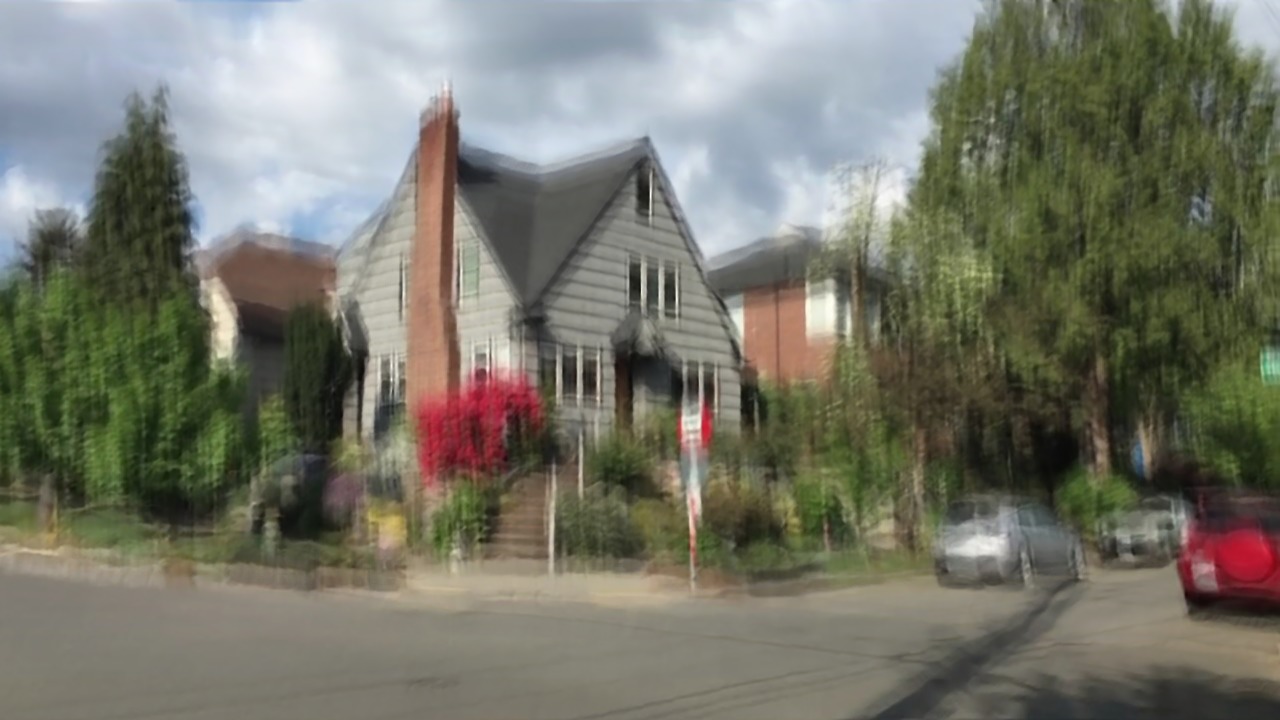} & 
				\includegraphics[bb=500 300 800 450,clip=True,width=\widthscalesix \textwidth]{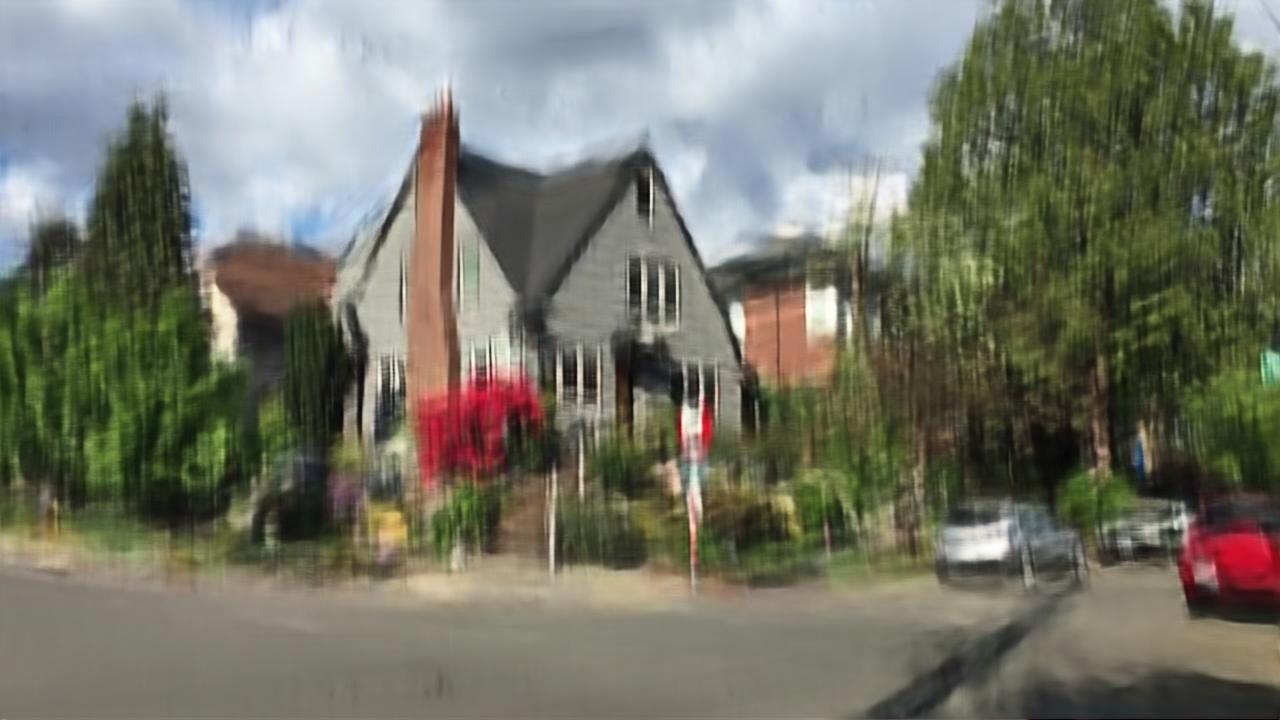} & 
								\includegraphics[bb=500 300 800 450,clip=True,width=\widthscalesix \textwidth]{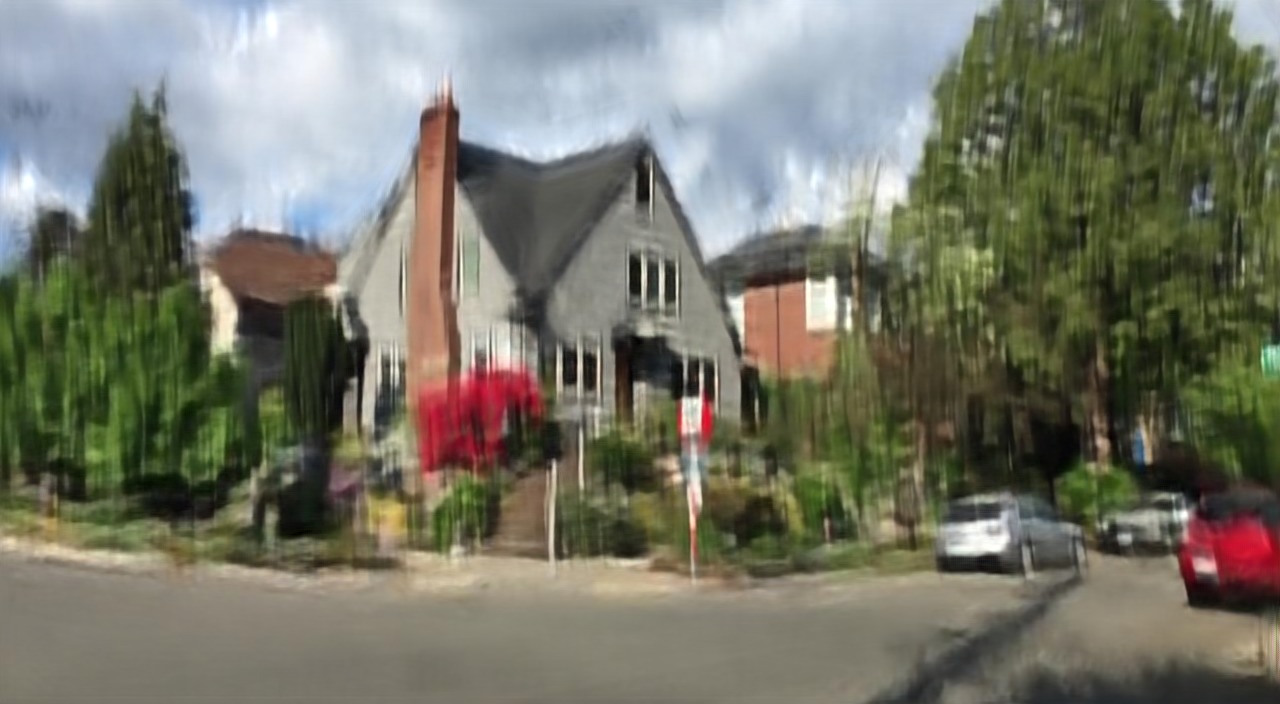} & 
				\includegraphics[bb=500 300 800 450,clip=True,width=\widthscalesix \textwidth]{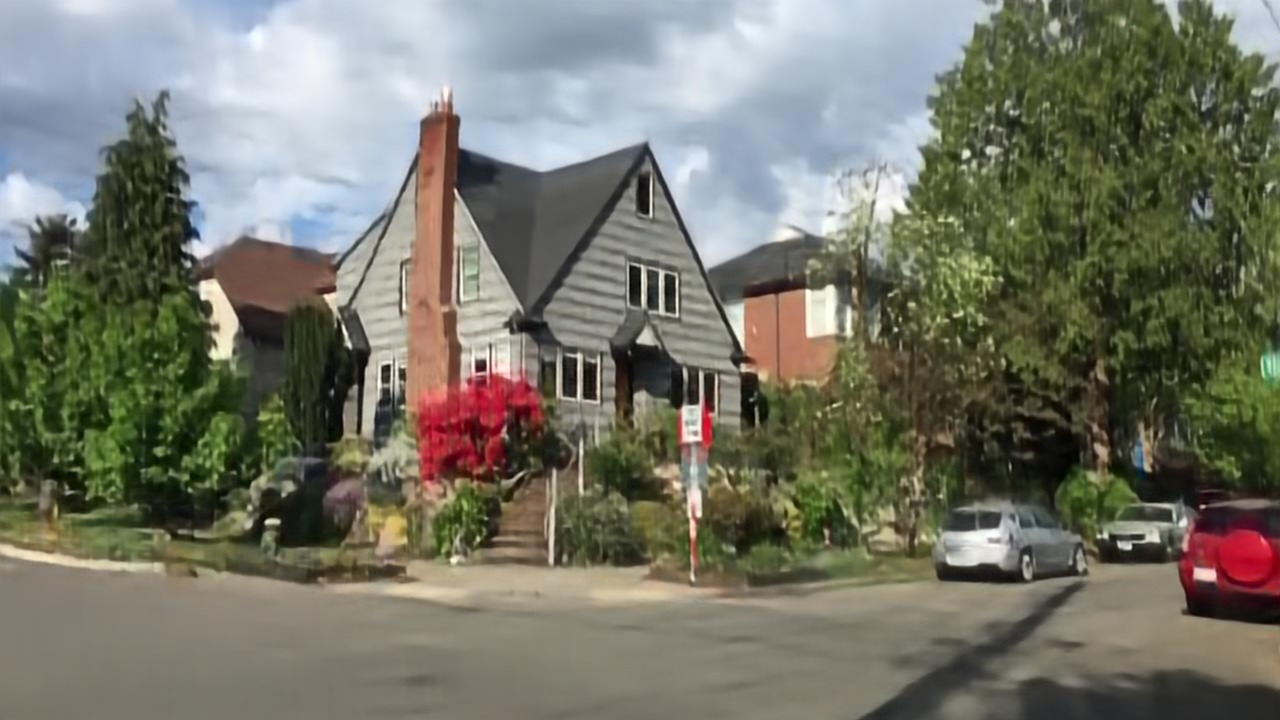}	
				\\
				\includegraphics[width=0.12\textwidth]{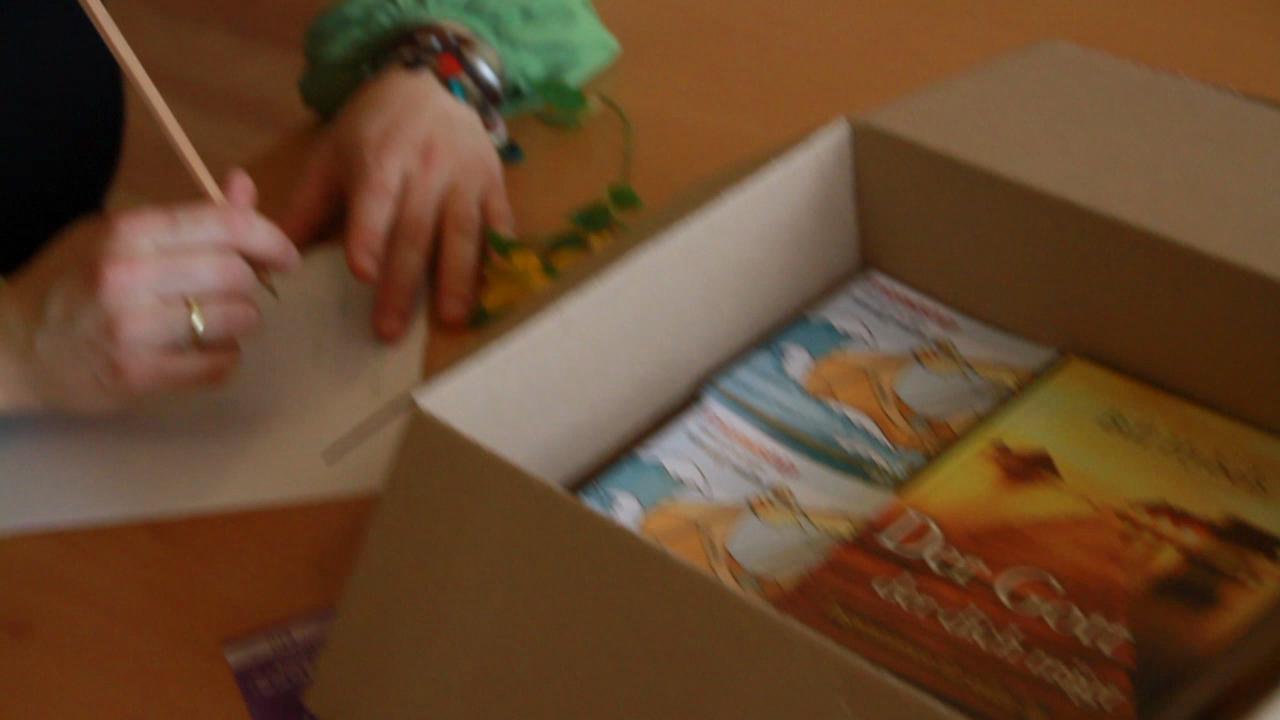}
				&
				\includegraphics[bb=800 20 1200 220,clip=True,width=\widthscalesix \textwidth]{video_real/00018_blur.jpg} & 
				\includegraphics[bb=800 20 1200 220,clip=True,width=\widthscalesix \textwidth]{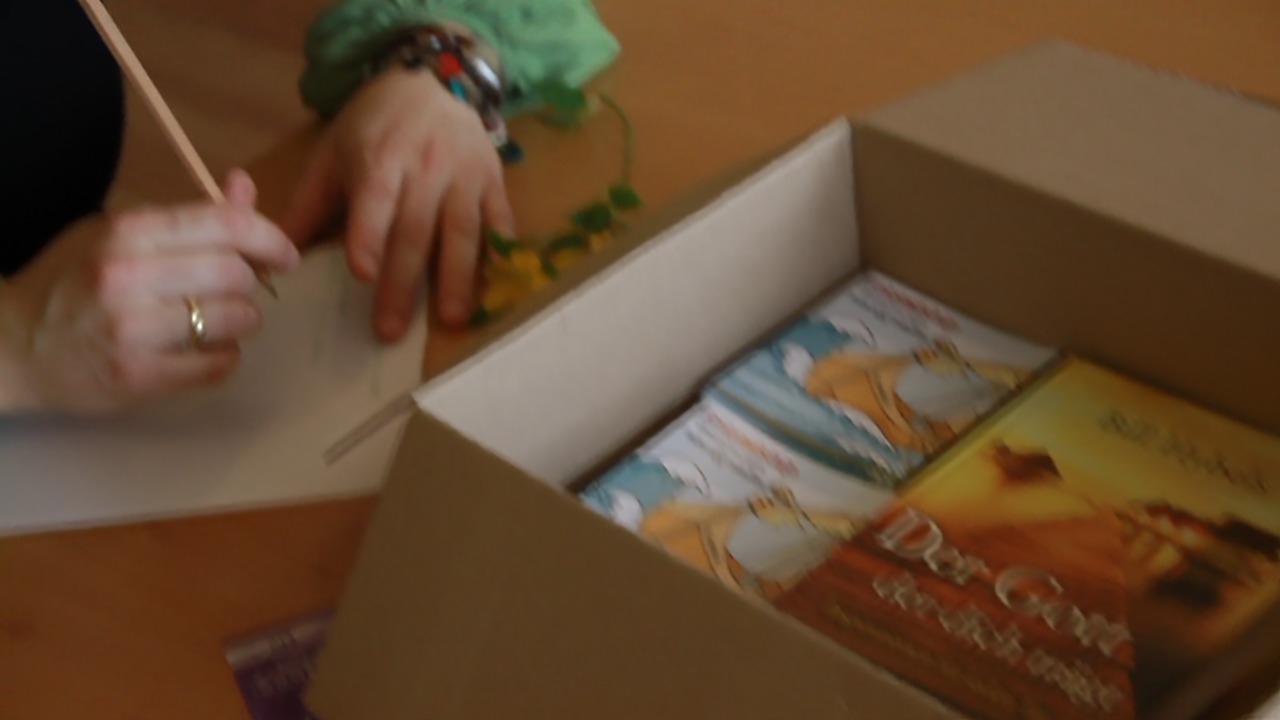} & 
				\includegraphics[bb=800 20 1200 220,clip=True,width=\widthscalesix \textwidth]{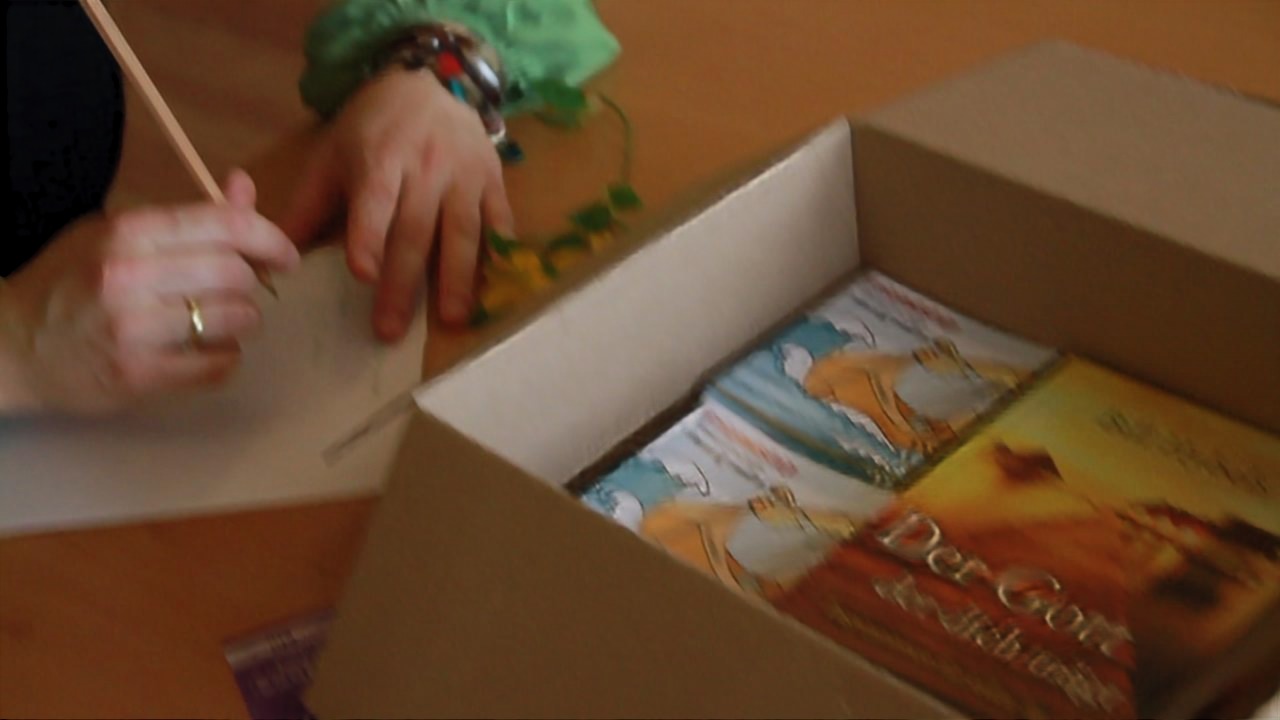} & 
								\includegraphics[bb=800 20 1200 220,clip=True,width=\widthscalesix \textwidth]{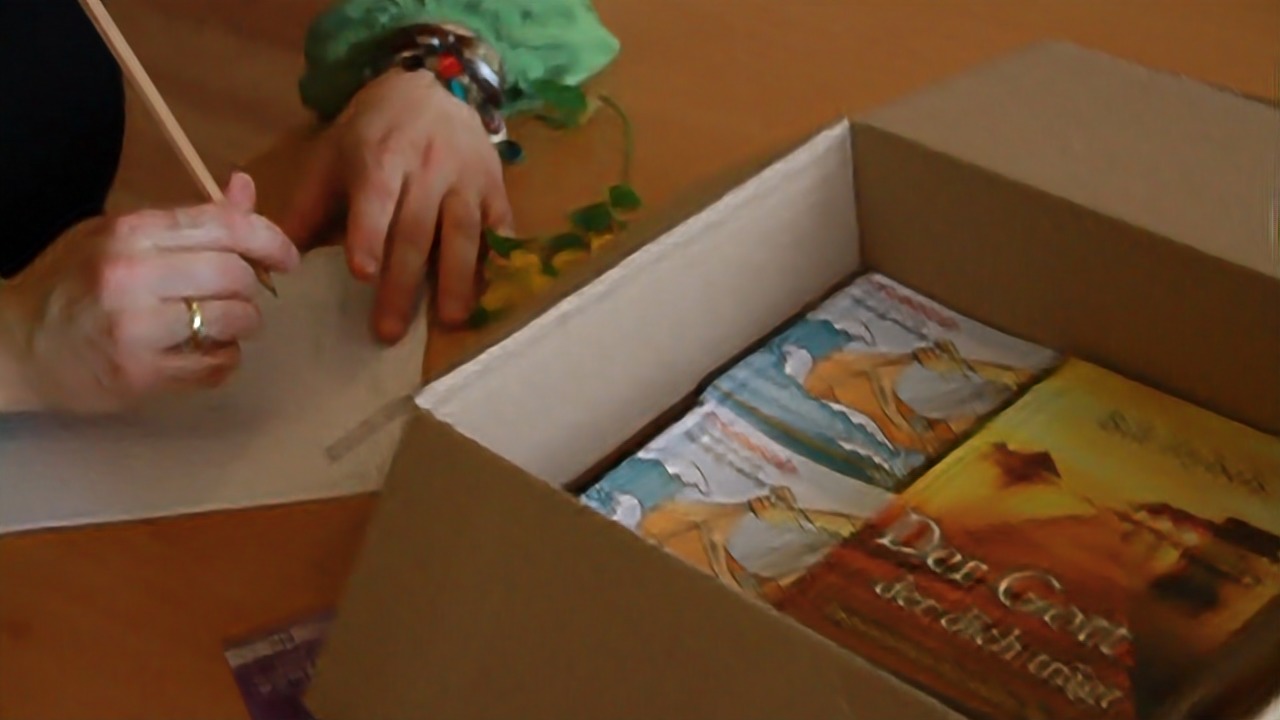} & 
				\includegraphics[bb=800 20 1200 220,clip=True,width=\widthscalesix \textwidth]{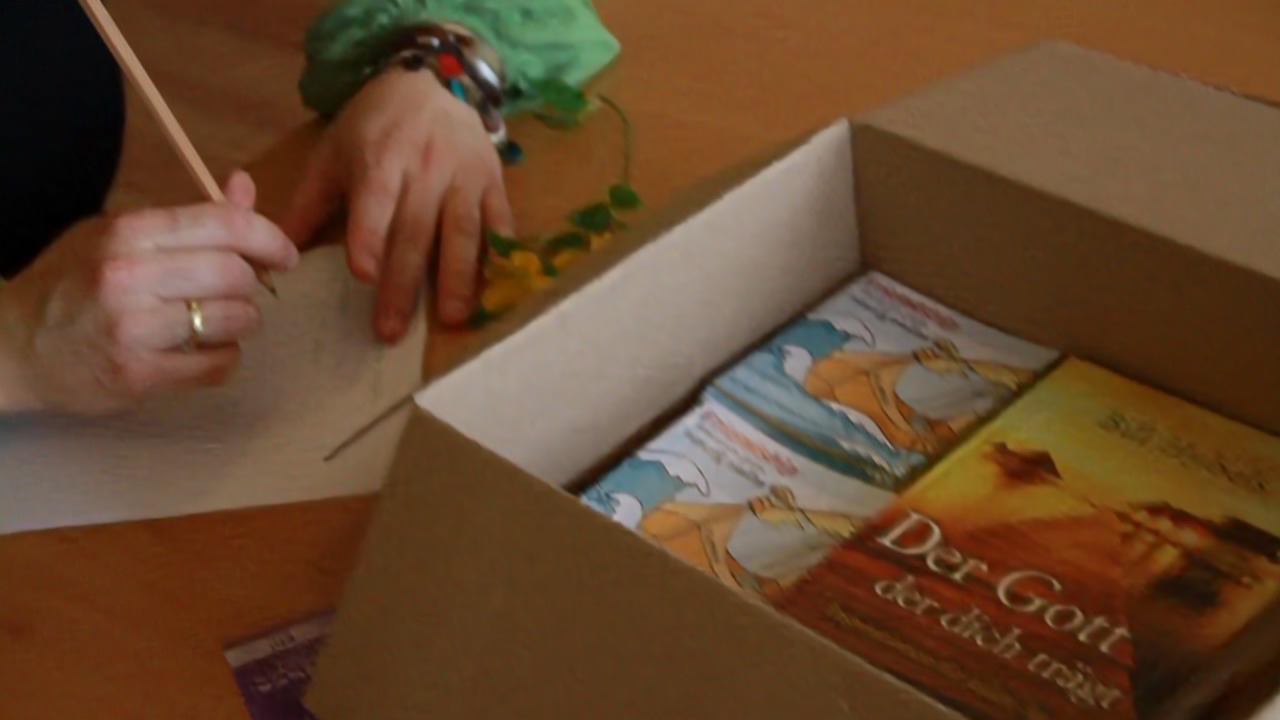}				\vspace{-3mm}

				\\ 
				\\ 
								(a) Blurred Image&

				(b) Blurred patch& 
				(c) Kim \etal~\cite{hyun2015generalized} & 
				(d) DVD~\cite{su2017deep} & 
				(e) OVD~\cite{hyun2017online}& 
				(f) Ours
				\\
	\end{tabular}
	\vspace{-0em}
	\caption{Visual comparisons of video deblurring results on two test frames from the DVD dataset~\cite{su2017deep}. Key blurred patches are shown in (b), while zoomed-in patches from the deblurred results are shown in (c)-(f). (best viewed in high resolution).}
\label{fig:video}
	\vspace{-0.0em}
\end{figure*}

\begin{table*}[!htbp]
\centering
\caption{Performance Comparison of our method with existing deblurring algorithms on single image deblurring benchmark dataset \cite{nah2017deep}.\label{TableGopro}}
\resizebox{17.5cm}{!}{

\begin{tabular}{|c|c|c|c|c|c|c|c|c|c|c|}
\hline
Method & Xu \cite{xu2013unnatural} & Whyte \cite{whyte2012non} & Kim \cite{hyun2013dynamic} & Sun\cite{sun2015learning} & MBMF \cite{gong2017motion} & MS-CNN \cite{nah2017deep} & DeblurGAN \cite{kupyn2017deblurgan} & SRN \cite{tao2018scale} & SVRNN \cite{zhang2018dynamic} & SARN (Ours) \\
\hline
PSNR (dB) & 21 & 24.6 & 23.64 & 24.64 &  26.4  & 29.08 & 28.7 & 30.26 & 29.19 & \textbf{31.13} \\
SSIM & 0.7407 & 0.8458 & 0.8239 & 0.843 & 0.8632 & 0.914 & 0.858 & 0.934 & 0.931 & \textbf{0.947}\\
Time (s) & 3800 & 700 & 3600 & 1500 & 1200 & 6 & 1 & 0.4 & 1 & \textbf{0.02} \\
Size (MB) & - & - & - & 54.1 & 41.2 & 55 & 50 & 28 & 37.1 & \textbf{11.2} \\
\hline
\end{tabular}
}
\end{table*}

\begin{table}[htbp]
\centering
\caption{\label{tab:vid_quantitative} Performance comparison of our method with existing video deblurring approaches on the benchmark dataset~\cite{su2017deep}.}
\resizebox{8.5cm}{!}{

\begin{tabular}{|c|c|c|c|c|c|c|}
\hline
Method & WFA & DVD~\cite{su2017deep} & DVD~\cite{su2017deep} & OVD & Ours-& Ours- \\
& \cite{delbracio2015hand} & Noalign& Flow& \cite{hyun2017online}& Multi&  Recurrent \\
\hline
PSNR (dB) & 28.35 & 30.05 & 30.05 & 29.95 & 30.60 &\textbf{31.15} \\
Time (s) & 15 & 0.7 & 5 & 0.3 &\textbf{0.02} & 0.05\\
Size (MB) &  - & 61.2 & 61.2 & \textbf{11.0} & 11.2 & 12.4\\
\hline
\end{tabular}
}
\end{table}

\section{Experimental Results}

In this section, we carry out quantitative and qualitative comparisons of our architectures with state-of-the-art methods for image as well as video deblurring tasks.

\subsection{Image Deblurring}
Due to the complexity of the blur present in general dynamic scenes, conventional uniform blur model based deblurring approaches struggle to perform well~\cite{nah2017deep}. However, we compare with conventional non-uniform deblurring approaches by Xu et al. \cite{xu2013unnatural} and Whyte et. al. \cite{whyte2012non} (proposed for static scenes)  and \cite{hyun2013dynamic} (proposed for dynamic scenes). Further, we compare with state-of-the-art end-to-end learning based methods \cite{nah2017deep,kupyn2017deblurgan,zhang2018dynamic,tao2018scale}. The source codes and trained models of competing methods are publicly available on the authors' websites, except for \cite{hyun2013dynamic} and \cite{zhang2018dynamic} whose results have been reported in previous works \cite{zhang2018dynamic,tao2018scale}. Public implementations with default parameters were used to obtain qualitative results on selected test images.

\noindent \textbf{Quantitative Evaluation}
Quantitative comparisons using PSNR and SSIM scores obtained on the GoPro testing set are presented in Table \ref{TableGopro}. Since traditional methods cannot model combined effects of general camera shake and object motion~\cite{xu2013unnatural,whyte2012non} or forward motion and depth variations~\cite{hyun2013dynamic}, they fail to faithfully restore most of the images in the test-set. The below par performance of \cite{sun2015learning,gong2017motion} can be attributed to the fact that they use synthetic and simplistic blur kernels to train their CNN and employ traditional deconvolution methods to estimate the sharp image, which severely limits their applicability to general dynamic scenes. On the other hand, the method of \cite{kupyn2017deblurgan} trains a network containing instance-normalization layers using a mixture of deep-feature losses and adversarial losses, but leads to suboptimal performance on images containing large blur. The methods~ \cite{nah2017deep,tao2018scale} use a multi-scale strategy to improve capability to handle large blur, but fail in challenging situations. One can note that the proposed SARN significantly outperforms all prior works, including the spatially varying RNN based approach~\cite{zhang2018dynamic}. As compared to the state-of-the-art \cite{tao2018scale}, our network offers an improvement of $\sim 0.9$ dB. 
 
\noindent \textbf{Qualitative Evaluation}
Visual comparisons on different dynamic and 3D scenes are given in Fig.~\ref{fig:dynamic}. It shows that results of prior works suffer from incomplete deblurring or ringing artifacts. In contrast, our network is able to restore scene details more faithfully due to its effectiveness in handling large dynamic blur and preserving sharpness. Importantly, our method fares significantly better in terms of model-size and inference-time ($70\%$ smaller and $20\times$ faster than the nearest competitor \cite{tao2018scale} on a single GPU). An additional advantage over \cite{xu2013unnatural,whyte2012non} is that our model waives-off the requirement of parameter tuning during test phase.

\subsection{Video Deblurring}

\noindent \textbf{Quantitative Evaluation}
To demonstrate the superiority of our model, we compare  the  performance of our network with that of state-of-the-art video deblurring approaches on 10 test videos from the benchmark \cite{su2017deep}. Specifically, we  compare  our  models with conventional model of \cite{delbracio2015hand}, two versions of DVD \cite{su2017deep}, and OVD \cite{hyun2017online}. Source codes of competing methods are publicly available on the authors' websites, except for \cite{delbracio2015hand} whose results have been reported in~\cite{su2017deep}. Table \ref{tab:vid_quantitative} shows quantitative comparisons between our method and competing methods. We also include a baseline `Ours-Multi', which refers to a version of our network which takes a stacks 5 consecutive blurred frames as input (configuration of DVD-Noalign). `Ours-recurrent' refers to our final network involving recurrence at frame as well as feature level. The results indicate that our method significantly outperforms prior methods ($\sim 1$ dB higher).
\\
\textbf{Qualitative Evaluation}
Fig. \ref{fig:video} contains visual comparisons with \cite{hyun2015generalized,su2017deep,hyun2017online} on different test frames from the qualitative and quantitative subsets of ~\cite{su2017deep} which suffer from complex blur due to large motion. Although traditional method \cite{hyun2015generalized} models pixel-level blur using optical flow as a cue, its fails to completely deblur many scenes due to its simplistic assumptions on the kernels and the image properties. Learning based methods \cite{su2017deep,hyun2017online} fare better than \cite{hyun2015generalized} in several cases but still lead to artifacts in deblurring due to their sub-optimal network design. Our method generates sharper results and faithfully restores the scenes, while yielding significant improvements on images affected with large blur.

\begin{table}[t]
\centering
\caption{Quantitative comparisons of different versions of our single image deblurring network on GoPro testset~\cite{nah2017deep}.
\label{TableAblationSingle}} 
\resizebox{6.0cm}{!}{
\begin{tabular}{|c|c|c|c|c|c|}
\hline
DRMs  & 0 & 3 & 6 & 6\\
SA & \checkmark & \xmark & \xmark & \checkmark \\
\hline 
PSNR (dB) & 30.64 & 30.69 & 31.05 & 31.13\\
Size (MB) & 10.7& 10.9 & 11.2 & 11.2 \\
\hline
\end{tabular}
}
	\vspace{-1.0em}
\end{table}

\vspace{-0.4cm}
\section{Conclusions}
	\vspace{-0.7em}

We proposed efficient image and video deblurring architectures composed of convolutional modules that enable spatially adaptive feature learning through filter transformations and feature attention over spatial domain, namely deformable residual module (DRM) and self-attentive (SA) module. The DRMs implicitly address the shifts responsible for the local blur in the input image, while the SA module non-locally connects spatially distributed blurred regions. Presence of these modules awards higher capacity to our compact network without any notable increase in model size. Our network's key strengths are large receptive field and spatially varying adaptive filter learning capability, whose effectiveness is also demonstrated for video deblurring through a recurrent extension of our network. Experiments on dynamic scene deblurring benchmarks showed that our approach performs favorably against prior art and facilitates real-time deblurring. We believe our spatially-aware design can be utilized for other image processing and vision tasks as well, and we shall explore them in the future.\\

Refined and complete version of this work appeared in AAAI 2020.

{\small
\bibliographystyle{ieee}
\bibliography{ref}
}

\end{document}